\pdfoutput=1
\documentclass[10pt,twocolumn,letterpaper]{article}

\usepackage{cvpr}
\usepackage{times}
\usepackage{graphicx}
\usepackage{amsmath}
\usepackage{amssymb}

\usepackage{bm}
\usepackage{bbm}
\usepackage{algpseudocode,algorithm,algorithmicx}
\usepackage{subcaption}

\usepackage{pgfplots}
\usepackage{tikz}
\usetikzlibrary{arrows.meta}
\usetikzlibrary{calc,intersections}
\usetikzlibrary{positioning}

\definecolor{cstblue}{HTML}{006699}
\definecolor{cstlightblue}{HTML}{66CCFF}
\definecolor{cstorange}{HTML}{CC3300}
\definecolor{cstgray}{HTML}{585858}
\pgfplotsset{colormap={bluered}{color(0)=(blue!85);
                                color(1)=(blue!40!cstorange!40!white);
                                color(2)=(cstorange!85)}}

\usepackage[pagebackref=true,breaklinks=true,
            letterpaper=true,colorlinks,bookmarks=false]{hyperref}

\cvprfinalcopy 

\def\cvprPaperID{3203} 

\ifcvprfinal\fi
\begin{document}

\title{End-to-end Learning of Deterministic Decision Trees}

\author{
  Thomas Hehn
  \quad\quad\quad\quad\quad
  Fred A. Hamprecht\\
  HCI/IWR, Heidelberg University\\
  {\tt\small \{thomas.hehn,fred.hamprecht\}@iwr.uni-heidelberg.de}
}

\maketitle

\begin{abstract}
   Conventional decision trees have a number of favorable properties, including
interpretability, a small computational footprint and the ability to learn from
little training data. 
However, they lack a key quality that has helped fuel the deep learning
revolution: that of being end-to-end trainable, and to learn from scratch
those features that best allow to solve a given supervised learning problem.
Recent work (Kontschieder~2015) has addressed this deficit, but at the cost of
losing a main attractive trait of decision trees: the fact that each sample is
routed along a small subset of tree nodes only.
We here propose a model and Expectation-Maximization training scheme for 
decision trees that are fully probabilistic at train time, but after a 
deterministic annealing process become deterministic at test time.
We also analyze the learned oblique split parameters on image datasets and 
show that Neural Networks can be trained at each split node.
In summary, we present the first end-to-end learning scheme for 
deterministic decision trees and present results on par with or superior to 
published standard oblique decision tree algorithms.

\end{abstract}

\section{Introduction}
  When selecting a supervised machine learning technique, we are led by multiple
and often conflicting criteria. These include: how accurate is the resulting
model? How much training data is needed to achieve a given level of accuracy?
How interpretable is the model? How big is the computational effort at train
time? And at test time? How well does the implementation map to the available
hardware?

These days, neural networks have superseded all other approaches in
terms of achievable accuracy of the predictions; but state of the art networks
are not easy to interpret, are fairly hungry for training data, often require
weeks of GPU training and have a computational and memory footprint that rules
out their use on small embedded devices. Decision trees achieve inferior
accuracy, but are fundamentally more frugal. 

Both neural networks and decision trees are composed of basic computational
units, the perceptrons and nodes, respectively. A crucial difference between
the two is that in a standard neural network, all units are being evaluated for
every input; while in a decision tree with \(I\) inner split nodes, only 
\(\mathcal{O}(\log I)\) split nodes are visited. That is, in a decision tree, a 
sample is routed along a
single path from the root to a leaf, with the path conditioned on the sample's
features. 

It is this sparsity of the sample-dependent computational graph  that piques
our interest in decision trees; but we also hope to profit from their ability
to learn their comparatively few parameters from a small training set, and
their relative interpretability. 

One hallmark of neural networks is their ability to learn a complex combination
of many elementary decisions jointly, by end-to-end training using
backpropagation.
This is a feature that has so far been missing in
deterministic decision trees, which are usually constructed greedily without
subsequent tuning. We here propose a mechanism to remedy this deficit. 

\subsection{Contributions} 
\begin{itemize}
\item We propose a decision tree whose internal nodes are
  probabilistic and hence differentiable at train time.
  As a consequence, we are able to train the
  internal nodes jointly in an end-to-end fashion. This is true
  for linear nodes, but the property is maintained
  for more complex nodes, such as small Convolutional Neural Networks (CNNs)
  (section~\ref{sec:METHsplits}).
\item We derive an expectation-maximization style algorithm for finding the optimal
  parameters in a split node (section~\ref{sec:METHEM}).
  We develop a probabilistic split criterion that generalizes the
  long-established information gain~\cite{Quinlan1986}. The proposed criterion
  is asymptotically identical
  to information gain in the limit of very steep non-linearities, but allows to
  better model class overlap in the vicinity of a split decision boundary
  (section~\ref{sec:METHprobtrees}).
\item We demonstrate good results by making
  the nodes deterministic at test time, sending each sample along a unique path
  of only $\mathcal{O}(\log I)$ out of the $I$ inner nodes in a tree.
  We evaluate the performance of the proposed method on the same datasets as used
  in related work~\cite{Norouzi2015} (section~\ref{sec:EXPaccuracy}) and find
  steeper learning curves with respect to tree depth, as well as higher
  overall accuracy. 
  We show the benefit of regularizing the spatial derivatives
  of learned features when samples are images or image patches (section~\ref{sec:EXPparamviz}).
  Finally, we report preliminary experiments with minimalistic trees with
  CNNs as split feature.
\end{itemize}

\section{Related work}
  Decision trees and decision tree ensembles, such as random forests
\cite{Breiman2001}, are widely used for computer vision
\cite{Criminisi2013} and have proven effective on
a variety of classification tasks \cite{Delgado2014}.
In order to improve their performance for a specific task, it is common
practice to engineer its features to a specific task
\cite{Gall2009,Kontschieder2013,Fua2005}.
Oblique linear and non-linear classifiers using more than one feature at a
time have been benchmarked in~\cite{Menze2011},
but the available algorithms are, in contrast to our approach, limited to 
binary classification problems.

There have been several attempts to train decision trees using gradient
optimization techniques for more complex split functions.
Similarly to our approach, \cite{Montillo2013} have successfully 
approximated information gain using a sigmoid function and a smoothness 
hyperparameter.
However, that approach does not allow joint optimization of an entire tree.

In \cite{Norouzi2015}, the authors also propose an algorithm for optimization
of an entire tree with a given structure.
They show a connection between optimizing oblique splits and structured
prediction with latent variables.
As a result, they formulate a convex-concave upper bound on the tree's
empirical loss.
In order to find an initial tree structure, the work also relies
on a greedy algorithm, which is based on the same upper bound approach
\cite{Norouzi2015CO2}.
Their method is restricted to linear splits and
relies on the kernel trick to introduce higher order split features
as opposed to our optimization, which allows more complex split features.

Other advances towards gradient-based decision tree optimization rely on
either fuzzy or probabilistic split functions 
\cite{Suarez1999,Kontschieder2015,Jordan1994}.
In contrast to our approach, the assignment of a single sample to the leaves
remains fuzzy, respectively probabilistic, during prediction.
Consequently, all leaves and paths need to be evaluated for every sample,
which annihilates the computational benefits of trees.

We build on reference~\cite{Kontschieder2015}, which is closest to our work.
The authors use sigmoid functions to model the probabilistic routes and
employ the same log-likelihood objective.
In contrast to their work, we derive the alternating optimization using the
Expectation-Maximization approach as in~\cite{Jordan1994EM} and aim for a
deterministic decision tree for prediction.
Also, they start from a random, but balanced tree, because their algorithm
does not learn the structure of the tree.

Finally, connections between neural networks and decision tree ensembles have
been examined.
In \cite{Sethi1990,Welbl2014} decision tree ensembles are cast to neural
networks, which enables gradient descent training.
As long as the structure of the trees is preserved, the optimized parameters 
of the neural network can also be mapped back to the random forest. 
Subsequently, \cite{Richmond2016} cast stacked decision 
forests to convolutional neural networks and found an approximate mapping 
back.
In \cite{Ioannou2016,McGill2017} several models of neural networks with 
separate, conditional data flows are discussed.

Our work builds on various ideas of previous work,
however, none of these algorithms learn deterministic decision trees
with arbitrary split functions in an end-to-end fashion.

\section{Methods} \label{sec:methods}
  Consider a classification problem with input space 
\(\mathcal{X} \subset \mathbb{R}^p\) and output space
\(\mathcal{Y} = \{1,...,K\}\).
The training set is defined as 
\(\{\bm{x}_1,...,\bm{x}_N\} = \mathcal{X}_t \subset \mathcal{X}\) with
corresponding classes
\(\{y_1,...,y_N\} = \mathcal{Y}_t \subset \mathcal{Y}\).
We propose training a probabilistic decision tree model, which becomes
deterministic at test time.
\subsection{Standard decision tree and notation} \label{sec:model}
  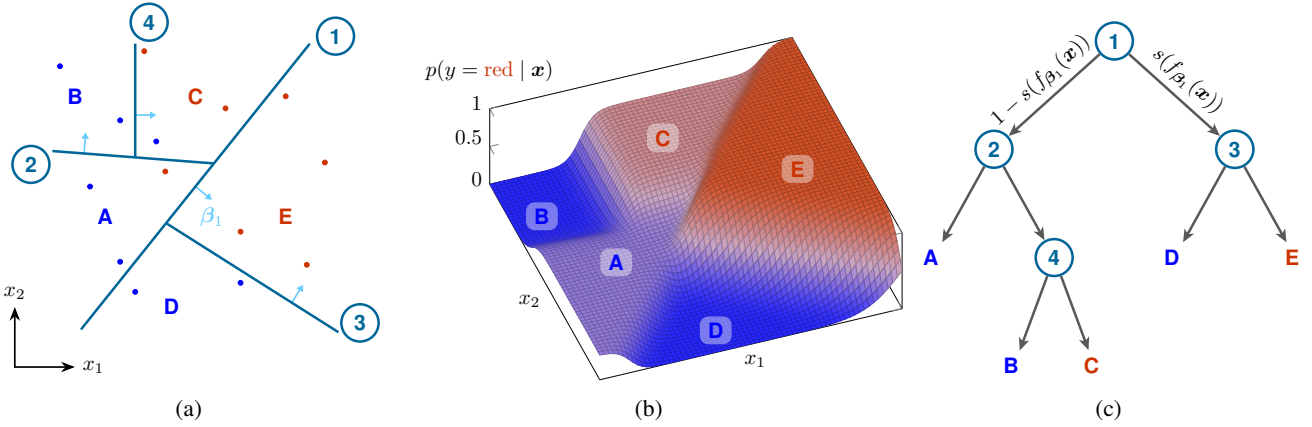
\begin{figure*}[t] 
    \tikzstyle{treenode}=[align=center, inner sep=3pt, text centered,
                          font=\sffamily]
    \tikzstyle{DN}=[treenode, circle, cstblue, font=\sffamily\bfseries, 
                    draw=cstblue, minimum width=1.5em, very thick]
    \tikzstyle{LN}=[treenode, font=\sffamily\bfseries,
                    minimum width=1.5em, very thick]
    \tikzstyle{LNr}=[LN, cstorange, fill=white, fill opacity=10]
    \tikzstyle{LNb}=[LN, blue     , fill=white, fill opacity=10]
    \tikzstyle{LNr3d}=[LNr, rounded corners, fill=white, fill opacity=0.4, text opacity=1.0]
    \tikzstyle{LNb3d}=[LNb, rounded corners, fill=white, fill opacity=0.4, text opacity=1.0]
    \begin{subfigure}[t]{0.30\textwidth} 
      \scalebox{0.8}{
        \begin{tikzpicture}[scale=5.0]
            \draw [blue, fill=blue]
              (0.15,0.50) circle (0.008);
            \draw [blue, fill=blue]
              (0.25,0.25) circle (0.008);
            \draw [cstorange, fill=cstorange]
              (0.4,0.55) circle (0.008);
            \draw [blue, fill=blue]
              (0.05,0.90) circle (0.008);
            \draw [blue, fill=blue]
              (0.25,0.72) circle (0.008);
            \draw [blue, fill=blue]
              (0.37,0.65) circle (0.008);
            \draw [cstorange, fill=cstorange]
              (0.33,0.95) circle (0.008);
            \draw [cstorange, fill=cstorange]
              (0.6,0.76) circle (0.008);
            \draw [blue, fill=blue]
              (0.30,0.15) circle (0.008);
            \draw [blue, fill=blue]
              (0.65,0.18) circle (0.008);
            \draw [cstorange, fill=cstorange]
              (0.8,0.80) circle (0.008);
            \draw [cstorange, fill=cstorange]
              (0.93,0.58) circle (0.008);
            \draw [cstorange, fill=cstorange]
              (0.65,0.35) circle (0.008);
            \draw [cstorange, fill=cstorange]
              (0.87,0.24) circle (0.008);
            \draw [{Stealth[]}-{Stealth[]},thick] 
                              (-0.1,0.1) node (yaxis) [above] {$x_2$} |- 
                              (0.1,-0.1) node (xaxis) [right] {$x_1$};
          \node at (0.1, 0.0) (hp1start) {};
          \node at (0.9, 1.0) (hp1end) {};
          \node at (0.0, 0.62) (hp2start) {};
          \node at (0.65, 0.57) (hp2end) {};
          \node at (1.0, 0.0) (hp3start) {};
          \node at (0.4, 0.38) (hp3end) {};
          \node at (0.3, 0.625) (hp4start) {};
          \node at (0.3, 1.0) (hp4end) {};
          \node at ($(hp1start)+0.5*(hp1end)-0.5*(hp1start)$) (nv1start) {};
          \node at ($(hp2start)+0.2*(hp2end)-0.2*(hp2start)$) (nv2start) {};
          \node at ($(hp3start)+0.3*(hp3end)-0.3*(hp3start)$) (nv3start) {};
          \node at ($(hp4start)+0.3*(hp4end)-0.3*(hp4start)$) (nv4start) {};
          \tikzstyle{NVec}=[-{Triangle[angle=60:1pt 3]},
                            thick,fill=cstlightblue,draw=cstlightblue]
          \tikzstyle{NVecNode}=[cstlightblue]
          \draw [NVec] 
            (nv1start.center) -- ($(nv1start.center)!0.07cm!-90:(hp1end)$)
            node [NVecNode, below] {$\bm{\beta}_1$};
          \draw [NVec] 
            (nv2start.center) -- ($(nv2start.center)!0.07cm!90:(hp2end)$);
          \draw [NVec] 
            (nv3start.center) -- ($(nv3start.center)!0.07cm!-90:(hp3end)$);
          \draw [NVec] 
            (nv4start.center) -- ($(nv4start.center)!0.07cm!-90:(hp4end)$);
          \draw [name path=hp1, very thick, draw=cstblue] (hp1start) -- (hp1end) 
            node[DN,right] {1};
          \draw [very thick, draw=cstblue] 
            (intersection of hp1start--hp1end and hp2start--hp2end) -- (hp2start)
            node[DN, below left] {2};
          \draw [very thick, draw=cstblue] 
            (intersection of hp1start--hp1end and hp3start--hp3end) -- (hp3start)
            node[DN, above right] {3};
          \draw [very thick, draw=cstblue] 
            (intersection of hp2start--hp2end and hp4start--hp4end) -- (hp4end)
            node[DN, above right] {4};
          \node [LNb] (LNA) at (0.2,0.4) {A};
          \node [LNb] (LNB) at (0.1,0.8) {B};
          \node [LNr] (LNC) at (0.5,0.8) {C};
          \node [LNb] (LND) at (0.42,0.1) {D};
          \node [LNr] (LNE) at (0.8,0.4) {E};
        \end{tikzpicture}
      } 
      \caption{}
      \label{fig:treefs}
    \end{subfigure}
    \begin{subfigure}[t]{0.39\textwidth} 
      \providecommand{\surfsamples}{65}
      \providecommand{\steepness}{45} 
      \providecommand{\xview}{-20}
      \providecommand{\yview}{70}
      \scalebox{0.8}{
        \providecommand{\surfsamples}{50}
\providecommand{\steepness}{50}
\providecommand{\xview}{-30}
\providecommand{\yview}{65}
\begin{tikzpicture}[declare function = {gamma=\steepness;}]
  \begin{axis}[
      xmajorticks=false,
      ymajorticks=false,
      xlabel=$x_1$, ylabel=$x_2$,
      zlabel={$p(y=\mathrm{\textcolor{cstorange}{red}}\mid\bm{x})$},
      zmin=0, zmax=1.0,
      point meta min=0,
      point meta max=1,
      view={\xview}{\yview},
      every axis z label/.style={ at={(ticklabel* cs:1.5)}},
  ]
  \addplot3[
    surf,
    domain=0:1,
    domain y=0:1,
    samples=\surfsamples,
  ] 
    {
      1.00 *   (1/(1+exp(-gamma*(x-0.8*y-0.1)))) 
           *   (1/(1+exp(-gamma*(0.38*x+0.6*y-0.38)))) 
    + 0.66 * (1-1/(1+exp(-gamma*(x-0.8*y-0.1)))) 
           *   (1/(1+exp(-gamma*(0.05*x+0.65*y-0.4)))) 
           *   (1/(1+exp(-gamma*(x-0.3)))) 
    + 0.33 * (1-1/(1+exp(-gamma*(x-0.8*y-0.1)))) 
           * (1-1/(1+exp(-gamma*(0.05*x+0.65*y-0.4)))) 
    };
  \node [LNb3d] (LNA) at (axis cs:0.2, 0.4, 0.33) {A};
  \node [LNb3d] (LNB) at (axis cs:0.1, 0.8, 0.0) {B};
  \node [LNr3d] (LNC) at (axis cs:0.5, 0.8, 0.66) {C};
  \node [LNb3d] (LND) at (axis cs:0.42, 0.1, 0.0) {D};
  \node [LNr3d] (LNE) at (axis cs:0.8, 0.4, 1.0) {E};
  \end{axis}
\end{tikzpicture}
      }
      \caption{}
      \label{fig:treeprobest3d}
    \end{subfigure}
    \begin{subfigure}[t]{0.30\textwidth} \centering
      \scalebox{0.8}{
        \begin{tikzpicture}[-{Stealth[length=6pt,width=5pt,inset=1.5pt]},
                            very thick, fill=cstgray, draw=cstgray, sloped, auto,
                            level/.style={sibling distance = 4cm/#1,
                                          level distance = 1.8cm}] 
          \node [DN] {1}
            child{ node [DN] {2} 
              child{ node [LNb] { A } } 
              child{ node [DN] {4}
                child{ node [LNb] { B } } 
                child{ node [LNr] { C } } 
              }
              edge from parent node[above] 
              {$1-s(f_{\bm{\beta}_1}(\bm{x}))$}
            }
            child{ node [DN] {3}
              child{ node [LNb] { D } } 
              child{ node [LNr] { E } } 
              edge from parent node[above] 
              {$s(f_{\bm{\beta}_1}(\bm{x}))$}
            };
        \end{tikzpicture}
      }
      \caption{}
      \label{fig:treediag}
    \end{subfigure}
    \caption{Probabilistic oblique decision trees.
             \subref{fig:treefs}) A feature space with a binary 
             classification problem tessellated by an example oblique decision tree.
             The oblique splits 
             (\textcolor{cstblue}{1}-\textcolor{cstblue}{4})
             partition the feature space into five different leaves (A-E).
             \subref{fig:treeprobest3d}) The predicted
             $p(y =\mathrm{\textcolor{cstorange}{red}} \mid \bm{x})$
             (eq.\ \ref{eq:probest}) of the oblique decision tree when a
             probabilistic split (eq.\ \ref{eq:sigmoid}) is used.
             \subref{fig:treediag}) The corresponding tree diagram.}
             \label{fig:treeprobest}
  \end{figure*}
  In binary decision trees (figure \ref{fig:treediag}),
  split functions
  \(s : \mathbb{R} \rightarrow [0,1]\) determine the routing of a sample
  through the tree, conditioned on that sample's features.
  The split function controls whether
  the splits are deterministic or probabilistic.
  The prediction is made by the leaf node that is reached by the sample.

  \textbf{Split nodes.}
  Each split node \(i \in \{1,...,I\}\) computes a split feature
  from a sample, and sends that feature to into a split function.
  That function is a map 
  \(f_{\bm{\beta}_i} : \mathbb{R}^p \rightarrow \mathbb{R}\)
  parametrized by \(\bm{\beta}_i\).
  For example, \emph{oblique splits} are a linear combination of the input as in
  \(f_{\bm{\beta}_i}(\bm{x}) = (\bm{x}^T, 1) \cdot \bm{\beta}_i \) with
  \(\bm{\beta}_i \in \mathbb{R}^{p+1}\).
  Similarly, an axis-aligned split perpendicular to axis \(a\) is represented
  by an oblique split whose only non-zero parameters are at index \(a\) and \(p+1\).
  We write \(\bm{\theta}_{\bm{\beta}} = (\bm{\beta}_1,...,\bm{\beta}_I)\) to
  denote the collection of all split parameters in the tree.
  
  \textbf{Leaf nodes.}
  Each leaf \(\ell \in \{1,...,L\}\) stores a categorical distribution over 
  classes \(k \in \{1,...,K\}\) in a vector
  \(\bm{\pi}_\ell \in [0,1]^K\).
  These vectors are normalized such that the probability of all classes in
  a leaf sum to \(\sum_{k=1}^K (\bm{\pi}_\ell)_k = 1\).
  We define \(\bm{\theta}_{\bm{\pi}} = (\bm{\pi}_1, ...,\bm{\pi}_L)\) to
  include all leaf parameters in the tree.
  
  \textbf{Paths.}
  For each leaf node there exists one unique set of split outcomes, called
  a path.
  We define the probability that a sample \(\bm{x}\) takes the path to leaf
  \(\ell\) as
  \begin{equation}
    \mu_\ell(\bm{x};s,\bm{\theta}_{\bm{\beta}}) = 
    \prod_{r \in \mathcal{R}_\ell} s(f_{\bm{\beta}_r}(\bm{x}))
    \prod_{l \in \mathcal{L}_\ell} \big(1- s(f_{\bm{\beta}_l}(\bm{x}))\big).
  \end{equation}
  Here, \(\mathcal{R}_\ell \subset \{1,...,I\}\) denotes the splits on the
  path which contain \(\ell\) in the right subtree.
  Analogously, \(\mathcal{L}_\ell \subset \{1,...,I\}\) denotes splits
  which contain \(\ell\) in the left subtree.
  In figure \ref{fig:treediag} this means that 
  \(\mathcal{R}_{\mathrm{B}} = \{2\}\) and
  \(\mathcal{L}_{\mathrm{B}} = \{1,4\}\).
  Also note that in the following, we will omit the \(s\) dependency
  whenever we do not consider a specific function.
  
  The prediction of the entire decision tree is given by multiplying the
  path probability with the corresponding leaf prediction:
  \begin{equation}
    p(y|\bm{x};\bm{\theta}) = 
    \sum_{\ell=1}^L 
    (\bm{\pi}_\ell)_y
    \mu_\ell(\bm{x};\bm{\theta}_{\bm{\beta}}).
    \label{eq:probest}
  \end{equation}
  Here, \(\bm{\theta} = (\bm{\theta}_{\bm{\beta}}, \bm{\theta}_{\bm{\pi}})\)
  comprises all parameters in the tree.
  This representation of a decision tree allows to choose between different
  split features and different split functions, by
  varying the functions \(f\) and \(s\), respectively.
  
  In standard deterministic decision trees as proposed in \cite{Breiman1984},
  the split function is
  a step function \(s(x) = \Theta(x)\) with
  \(\Theta(x) = 1\) if \(x > 0\) and \(\Theta(x) = 0\) otherwise.

\subsection{Probabilistic decision tree} \label{sec:METHprobtrees}
  We now introduce a defining characteristic of the proposed method.
  Rather than sending a sample deterministically down the left or right
  subtree, depending on its features \(x\), we send it left or right with a
  probability
  \begin{equation}
    s(f(x)) = \sigma(f(x)) = \frac{1}{1+e^{-f(x)}}.
    \label{eq:sigmoid}
  \end{equation}
  This corresponds to regarding each split in the tree as a Bernoulli decision
  with mean \(\sigma(f(x))\) and as a result equation~\ref{eq:probest} is the
  expected value over the possible outcomes.
  Figure \ref{fig:treeprobest3d} shows the prediction from equation 
  \ref{eq:probest} in the probabilistic
  case for a class \(y =\) ``red'' on the classification problem illustrated
  in figure \ref{fig:treefs}.
  
  To train our probabilistic decision trees, we choose as objective the
  maximization of the empirical log-likelihood of the training
  data:
  \begin{equation}
    \max_{\bm{\theta}}
    \pounds(\bm{\theta};\mathcal{X}_t,\mathcal{Y}_t)
    =
    \max_{\bm{\theta}}
    \sum_{n = 1}^N \log p(y_n|\bm{x}_n;\bm{\theta}).
    \label{eq:objective}
  \end{equation}
  
  Importantly, while we propose to use a probabilistic decision tree for
  training, we use a deterministic decision tree for prediction.
  To better match the models used at train and test time, we introduce a
  hyperparameter \(\gamma\), which 
  steers the steepness of the split function by scaling the split
  feature~\cite{Montillo2013}
  \begin{equation}
    s(f(x))
    = \sigma_\gamma(f(x))
    = \sigma(\gamma f(x)).
    \label{eq:gamma}
  \end{equation}
  Note, for \(\gamma\to\infty\) the model resembles a deterministic decision tree, since
  \(\sigma_\infty(f(x)) = \Theta(f(x))\).
  During training, we iteratively increase \(\gamma\), akin a temperature
  cooling schedule in deterministic annealing~\cite{Rose1990}.

\subsection{Expectation-Maximization} \label{sec:METHEM}
  For the optimization of the log-likelihood (equation \ref{eq:objective}), 
  we propose a gradient-based, 
  EM-style optimization strategy,
  which requires
  \(f\) and \(s\) to be differentiable with respect to the split
  parameters \(\bm{\beta}_i\).
  The derivation of the EM-algorithm for this model follows the spirit 
  of~\cite{Jordan1994EM}.
  We introduce additional latent random variables \(z_{n,\ell}\), which 
  indicate that leaf \(\ell\) generated the class label of a given data point
  \(\bm{x}_n\).
  Including these latent variables, the optimization objective 
  (eq.\ \ref{eq:objective}) becomes the complete-data log-likelihood (including latent 
  variables)
  \begin{equation}
    \pounds(\bm{\theta};\mathcal{X}_t,\mathcal{Y}_t,\mathcal{Z}_t) = 
    \sum_{n = 1}^N \sum_{\ell = 1}^L 
    z_{n,\ell} \log \big( 
    (\bm{\pi}_\ell)_{y_n}
    \mu_\ell(\bm{x}_n;\bm{\theta}_{\bm{\beta}}) \big).
  \end{equation}

  \textbf{E-Step.}
  In the Expectation-Step, the expected value of the complete-data log-likelihood
  over the latent variables given the previous parameters \(\bm{\theta}'\)
  is computed
  \begin{equation}
    Q(\bm{\theta}|\bm{\theta}') = 
    E_{\mathcal{Z}_t\mid\mathcal{X}_t,\mathcal{Y}_t;\bm{\theta}'}[
    \pounds(\bm{\theta};\mathcal{X}_t,\mathcal{Y}_t,\mathcal{Z}_t)].
  \end{equation}
  For this purpose, it is necessary to compute the probability that
  \(z_{n,\ell} = 1\) for each training sample \(n\):
  \begin{align}
    \hspace{-0.1cm}
    h_{n,\ell} :=
    &~p(z_{n,\ell} = 1 \mid \bm{x}_n, y_n; \bm{\theta}') \\
    =&~
    \frac{p(y_n\mid z_{n,\ell} = 1, \bm{x}_n; \bm{\theta}') 
          p(z_{n,\ell} = 1 \mid \bm{x}_n; \bm{\theta}')}
         {p(y_n \mid \bm{x}_n; \bm{\theta}')} \\
    =&~
    \frac{(\bm{\pi}'_\ell)_{y_n}
          \mu_{\ell}(\bm{x}_n;\bm{\theta}'_{\bm{\beta}})}
         {\sum_{\ell' = 1}^L (\bm{\pi}'_{\ell'})_{y_n}
          \mu_{\ell'}(\bm{x}_n;\bm{\theta}'_{\bm{\beta}})}.
    \label{eq:EMhidden}
  \end{align}
  Thus, the expectation value of the complete-data log-likelihood yields
  \begin{equation}
    Q(\bm{\theta}|\bm{\theta}') = 
    \sum_{n = 1}^N \sum_{\ell = 1}^L 
    h_{n,\ell} \log \big( 
    (\bm{\pi}_\ell)_{y_n}
    \mu_\ell(\bm{x}_n;\bm{\theta}_{\bm{\beta}}) \big).
    \label{eq:Q}
  \end{equation}

  \textbf{M-Step.}
  In the Maximization-Step of the EM-Algorithm, the expectation value
  computed in the E-Step (eq.\ \ref{eq:Q}) is maximized to find updated
  parameters
  \begin{equation}
    \max_{\bm{\theta}} Q(\bm{\theta}|\bm{\theta}').
  \end{equation}
  Due to the latent variables we introduced, it is now possible to separate
  the parameter dependencies in the logarithm into a sum.
  As a result, the leaf predictions and split parameters are optimized
  separately.
  The optimization of the leaf predictions including the normalization
  constraint can be computed directly as
  \begin{equation}
    (\bm{\pi}_\ell)_k = \frac{\sum_{n = 1}^N \mathbbm{1}(y_n = k) h_{n,\ell}}
                             {\sum_{n = 1}^N h_{n,\ell}}.
    \label{eq:EMleaves}
  \end{equation}
  Here, the indicator
  function \(\mathbbm{1}(y_n = k)\) equals \(1\) if \(y_n = k\) and
  \(0\) otherwise.
  \\
  The optimization of the split parameters in the M-Step is performed using
  gradient based optimization.
  The separated objective for the split parameters without the 
  leaf predictions is
  \begin{equation}
    \max_{\bm{\theta}_{\bm{\beta}}}
    \sum_{n = 1}^N \sum_{\ell = 1}^L 
    h_{n,\ell} \log 
    \mu_\ell(\bm{x}_n;\bm{\theta}_{\bm{\beta}}).
    \label{eq:EMsplitobjective}
  \end{equation}
  We use the first-order gradient-based stochastic optimization
  Adam~\cite{Kingma2015} for optimization of the split parameters.
  
  In summary, each iteration of the algorithm requires evaluation
  of equations \ref{eq:EMhidden} and \ref{eq:EMleaves}, as well as at least
  one update of the split parameters based on equation
  \ref{eq:EMsplitobjective}.
  This iterative algorithm can be applied to a binary decision tree of any
  given structure.

\subsection{Complex splits and spatial regularization} \label{sec:METHsplits}
  The proposed optimization procedure only requires the split
  features \(f\) to be differentiable with respect to the split parameters.
  As a result, it is possible to implement more complex splits than axis-aligned
  or oblique splits.
  For example, it is possible to use a small Convolutional Neural Network (CNN)
  as split feature extractor for \(f\) and learn its parameters 
  (section \ref{sec:EXPCNN}).
  
  Furthermore, the optimization objective can also include regularization
  constraints on the parameters.
  This is useful to avoid overfitting and learn more robust patterns.
  When the inputs are from images, spatial regularization also reveals
  more discernible spatial structures in the learned parameters without
  sacrificing accuracy (section \ref{sec:EXPparamviz}).
  To encourage the learning of coherent spatial patterns at each
  split, we introduce a spatial regularization term
  \begin{equation}
    - \lambda \sum_{i=1}^I \bm{\beta}_i^T M \bm{\beta}_i
    \label{eq:regularization}
  \end{equation}
  to the maximization
  objective of the split features (eq.\ \ref{eq:EMsplitobjective})
  \cite{Eilers1996}.
  The matrix \(M\) denotes the Laplacian matrix when interpreting the image as
  a grid graph.
  For a single pixel, corresponding to weight \(\beta_i\), the diagonal element
  \(M_{ii}\)
  contains the number of neighboring pixels.
  If pixels \(i\) and \(j\) are neighboring pixels, then \(M_{ij} = M_{ji} = -1\).
  All remaining elements in \(M\) are 0.
  This regularization term penalizes spatial finite differences,
  encouraging similar parameters for neighboring pixels.
  The hyperparameter \(\lambda\) controls the regularization strength, with
  higher \(\lambda\) leads to stronger regularization.
  
\subsection{Structure learning}
  The foregoing shows how to fit a decision tree to training data, given the
  tree topology (parameter learning).
  We now turn to the learning of the tree itself (structure learning).
  We recommend, and evaluate in section~\ref{sec:EXPaccuracy}, a greedy 
  strategy: 
  Starting at the root, each split is considered and trained as a tree stump,
  consisting of one split and two leaf nodes.
  
  Since there are only two leaves, the log-likelihood objective 
  (eq.~\ref{eq:objective}) then resembles an approximation of the
  widely popular information gain criterion \cite{Quinlan1986,Quinlan1993}
  (section~\ref{sec:METHinfogain}).
  The previously found splits, in the more shallow levels of the tree, 
  deterministically route data to the split currently being trained.
  In particular this means that, at first, the root split is trained on the
  entire training data.
  After training of the first split, both leaves are discarded and replaced
  by new splits.
  According to the root split, the training data is deterministically divided into
  two subsets, which are now used to train the corresponding child nodes.
  This procedure is repeated until, some stopping criterion, \eg maximum
  depth, maximum number of leaves or leaf purity, is reached.
  After this greedy structure learning, the nodes in the entire resulting tree
  can be finetuned jointly as described in section~\ref{sec:METHEM}, this time with 
  probabilistic routing of all training data. 

\subsection{Relation to information gain and leaf entropies} \label{sec:METHinfogain}
  We now show that maximization of the log-likelihood of the probabilistic
  decision tree model approximately minimizes the weighted entropies in the
  leaves.
  The steeper the splits become, the better the approximation.
  
  To establish this connection we use hyperparameter \(\gamma\) to
  control the steepness of the probabilistic split function (eq.~\ref{eq:gamma}).
  We introduce the function \(\ell(\bm{x})\) that returns the index of the
  leaf sample \(\bm{x}\) reaches when the path is evaluated deterministically
  \begin{equation}
    \ell(\bm{x})
    = \sum_{\ell = 1}^L \ell \, \lim_{\gamma\to\infty}
    \mu_\ell(\bm{x};\sigma_\gamma,\bm{\theta}_{\bm{\beta}}).
  \end{equation}
  This simplifies the log-likelihood objective (eq.~\ref{eq:objective}) to
  \begin{equation}
    \max_{\bm{\theta}}
    \sum_{n = 1}^N \log 
    (\bm{\pi}_{\ell(\bm{x}_n)})_{y_n}
  \end{equation}
  because each sample reaches only one leaf.
  Let \(N_{\ell,k}\) be the number of training samples in leaf \(\ell\) with
  class \(k\) and \(N_{\ell} = \sum_{k=1}^K N_{\ell,k}\) denote all training
  samples in leaf \(\ell\).
  Since training samples with the same class and in the same leaf contribute the same
  term, the equations may be rearranged to
  \begin{equation}
    \max_{\bm{\theta}}
    \sum_{\ell = 1}^L
    \sum_{k = 1}^K 
    N_{\ell,k} \log 
    (\bm{\pi}_{\ell})_k.
  \end{equation}
  With \(\gamma\to\infty\), the optimal leaf predictions are the same as in
  a standard, deterministic decision tree, \ie
  \((\bm{\pi}_\ell)_k = \frac{N_{\ell,k}}{N_{\ell}}\).
  Accordingly, the objective can be rewritten as
  \begin{equation}
    \max_{\bm{\theta}}
    \lim_{\gamma\to\infty}
    \pounds(\bm{\theta};\mathcal{X}_t,\mathcal{Y}_t)
    =
    \min_{\bm{\theta}}
    \sum_{\ell = 1}^L
    \frac{N_{\ell}}{N}
    H_\ell.
  \end{equation}
  Here, 
  \(H_\ell = - \sum_{k = 1}^K
    (\bm{\pi}_{\ell})_k
    \log 
    (\bm{\pi}_{\ell})_k \)
  denotes the entropy in leaf \(\ell\).
  
  In conclusion, we have shown that for \(\gamma\to\infty\), maximizing the
  log-likelihood objective minimizes a weighted sum of leaf entropies.
  For the special case of a single split with two leaves, this is the
  same as maximizing the information gain.
  Consequently, the log-likelihood objective (eq.\ \ref{eq:objective}) can be 
  regarded as a generalization of the information gain criterion
  \cite{Quinlan1986} to an entire tree.

\section{Experiments}
  \begin{figure*}[t]
  \centering
  \begin{tikzpicture}
    \node (MNISTtest)  {\includegraphics[width=0.24\linewidth]{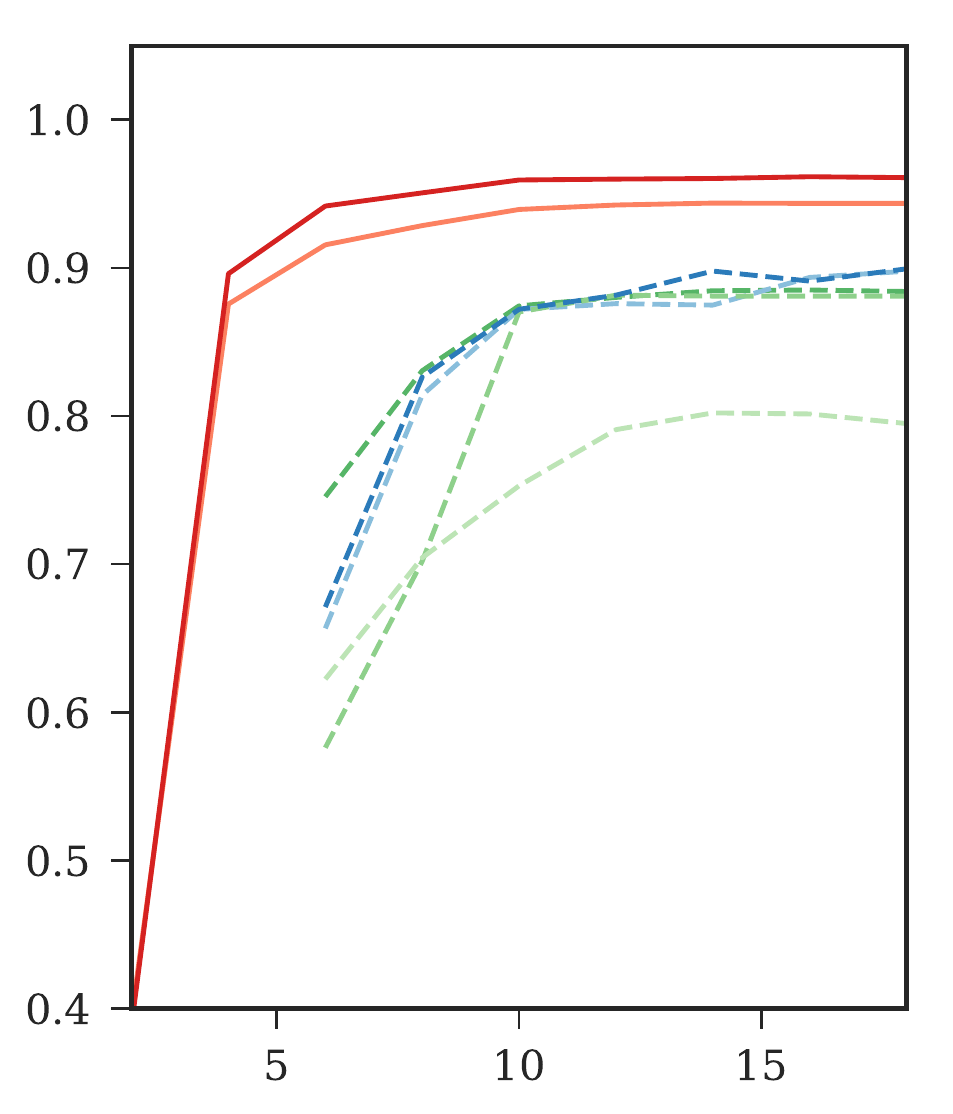}};
    \node[right=of MNISTtest, node distance=0cm, xshift=-1.2cm]
          (SensITtest)  {\includegraphics[width=0.24\linewidth]{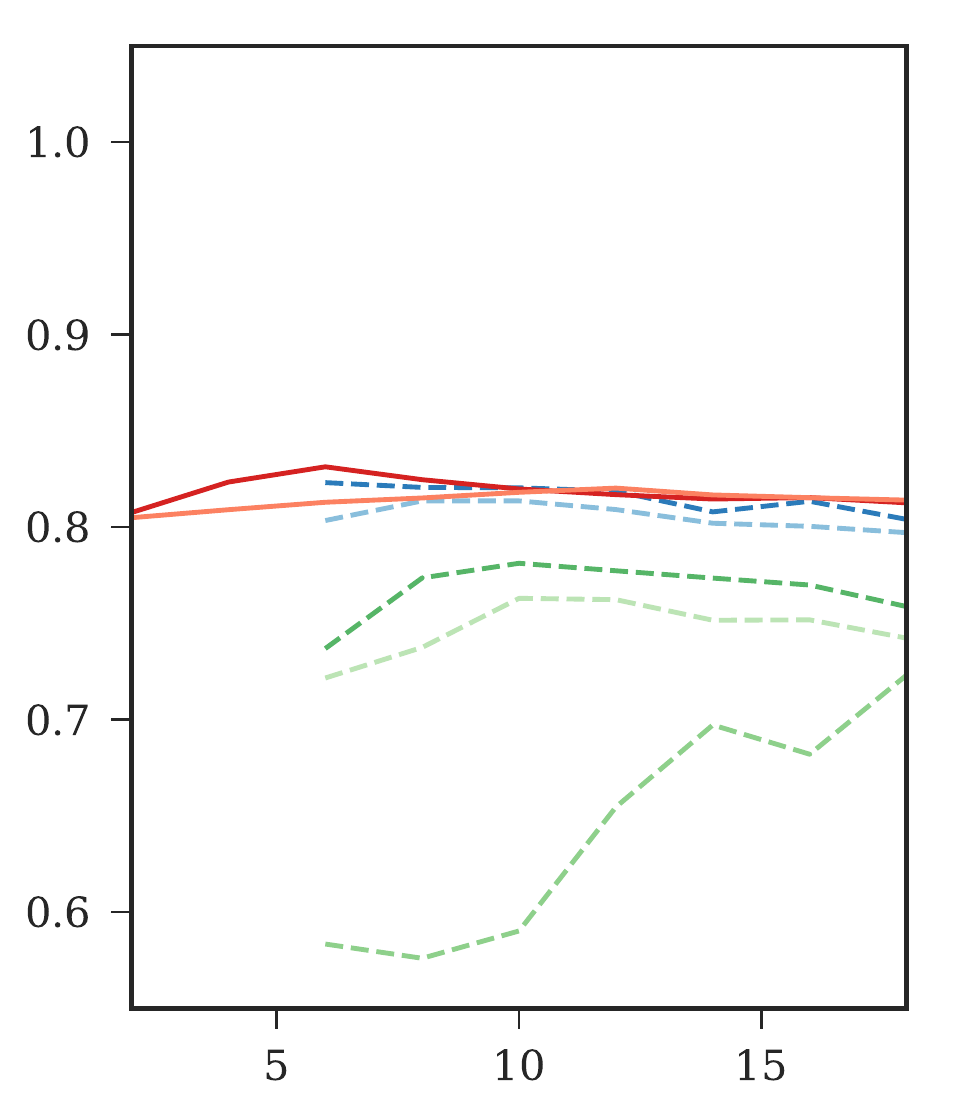}};
    \node[right=of SensITtest, node distance=0cm, xshift=-1.2cm]
          (Connect4test)  {\includegraphics[width=0.24\linewidth]{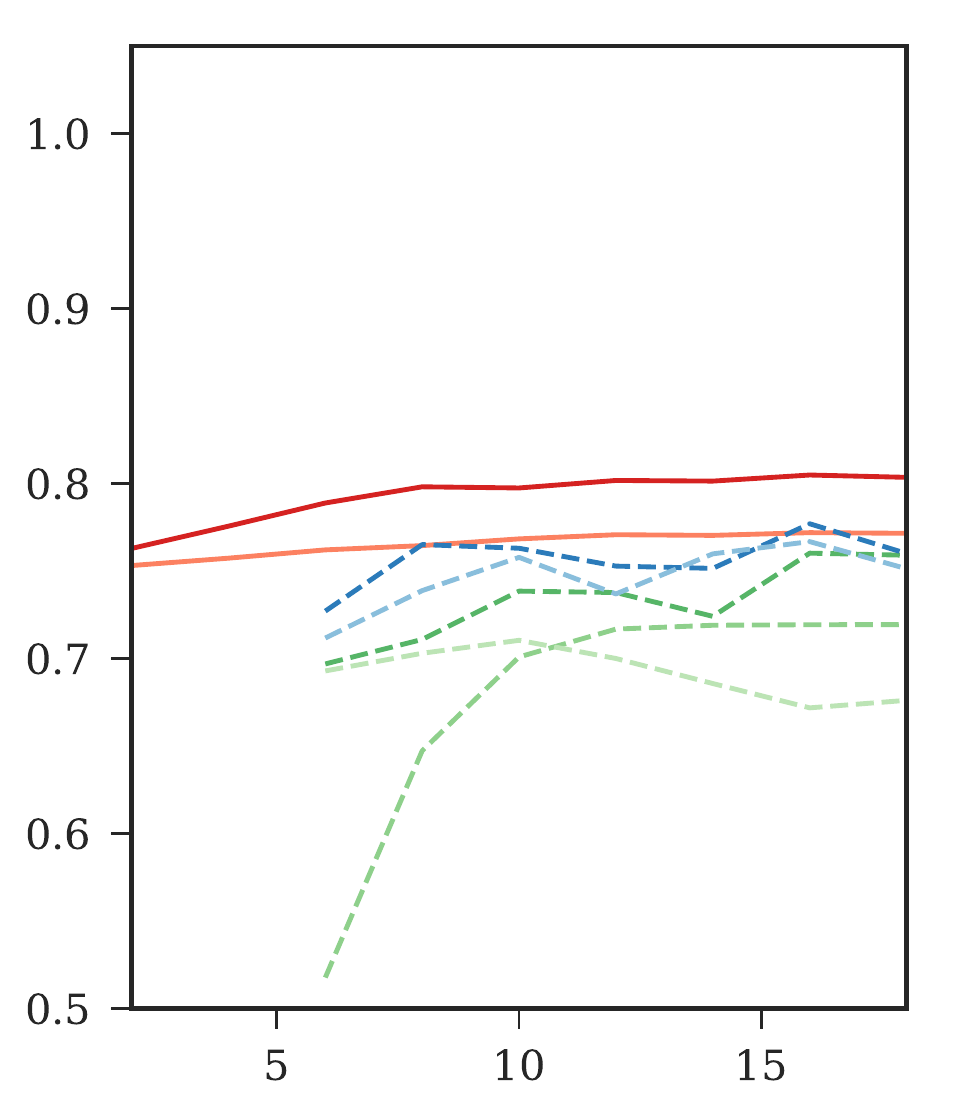}};
    \node[right=of Connect4test, node distance=0cm, xshift=-1.2cm]
          (Proteintest)  {\includegraphics[width=0.24\linewidth]{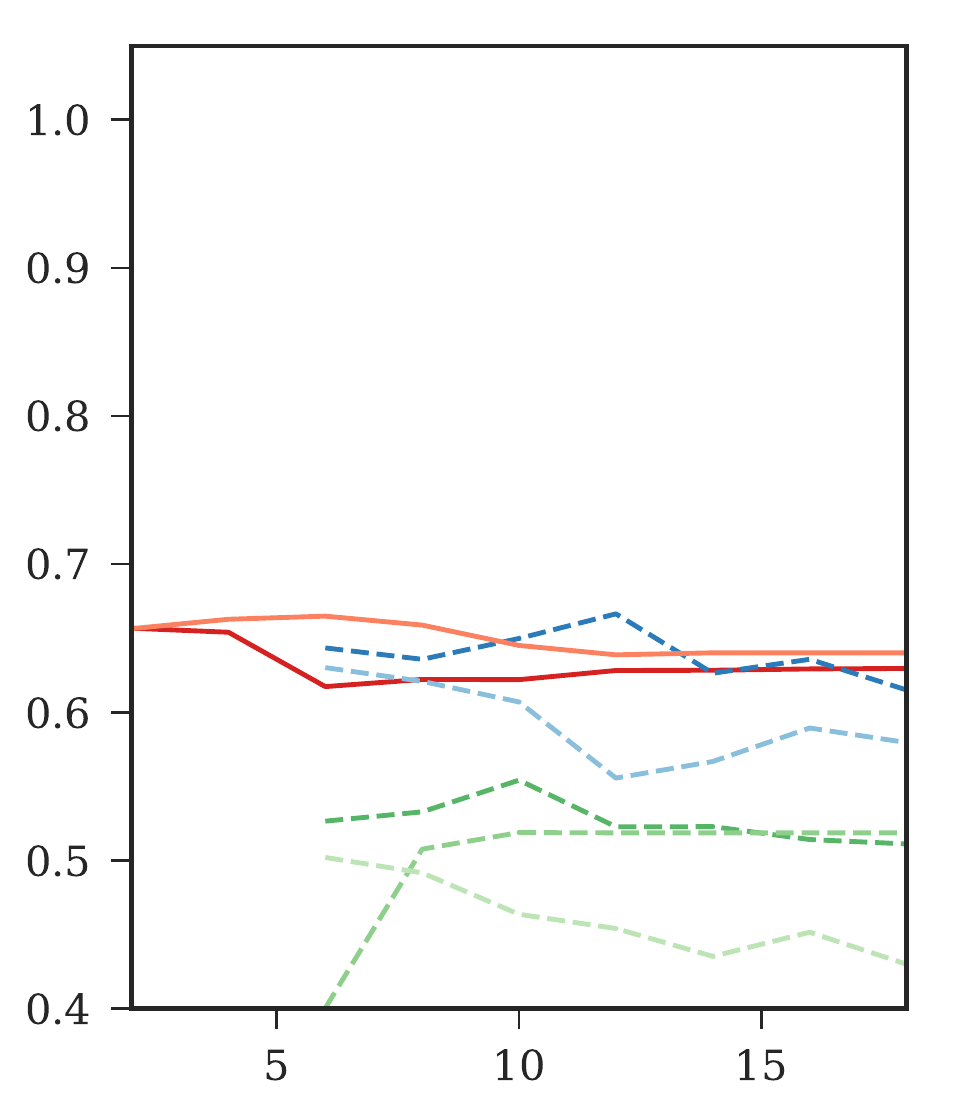}};
    \node[below=of MNISTtest, node distance=0cm, yshift=1.2cm] (MNISTtrain)  {\includegraphics[width=0.24\linewidth]{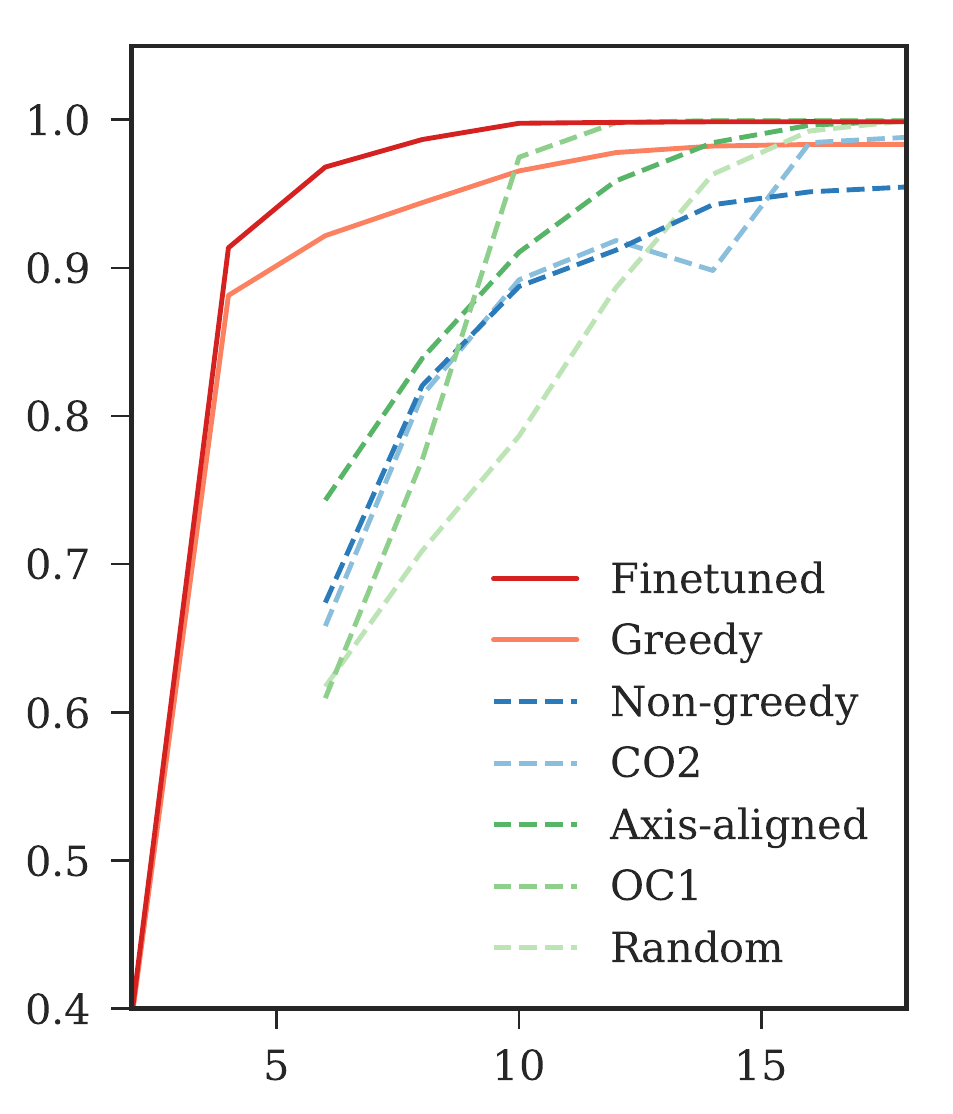}};
    \node[right=of MNISTtrain, node distance=0cm, xshift=-1.2cm]
          (SensITtrain)  {\includegraphics[width=0.24\linewidth]{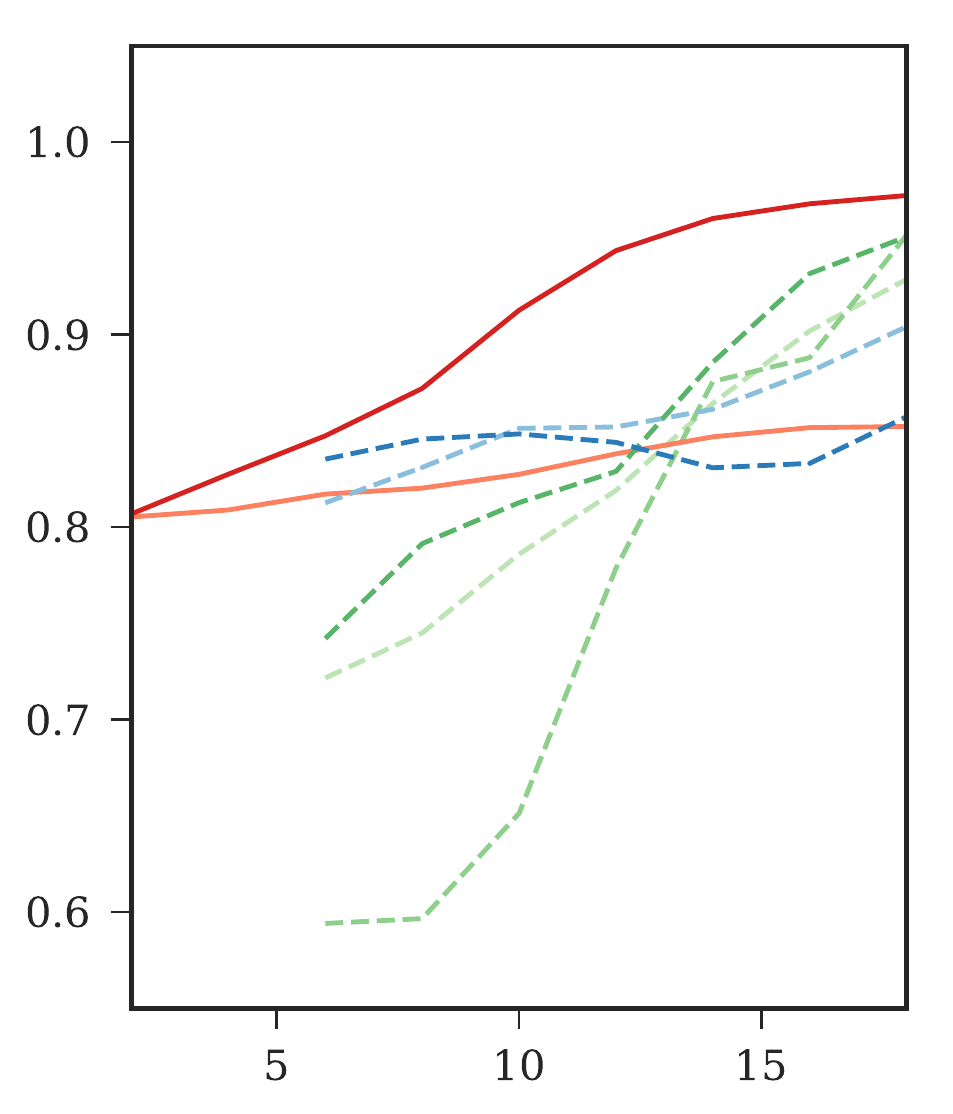}};
    \node[right=of SensITtrain, node distance=0cm, xshift=-1.2cm]
          (Connect4train)  {\includegraphics[width=0.24\linewidth]{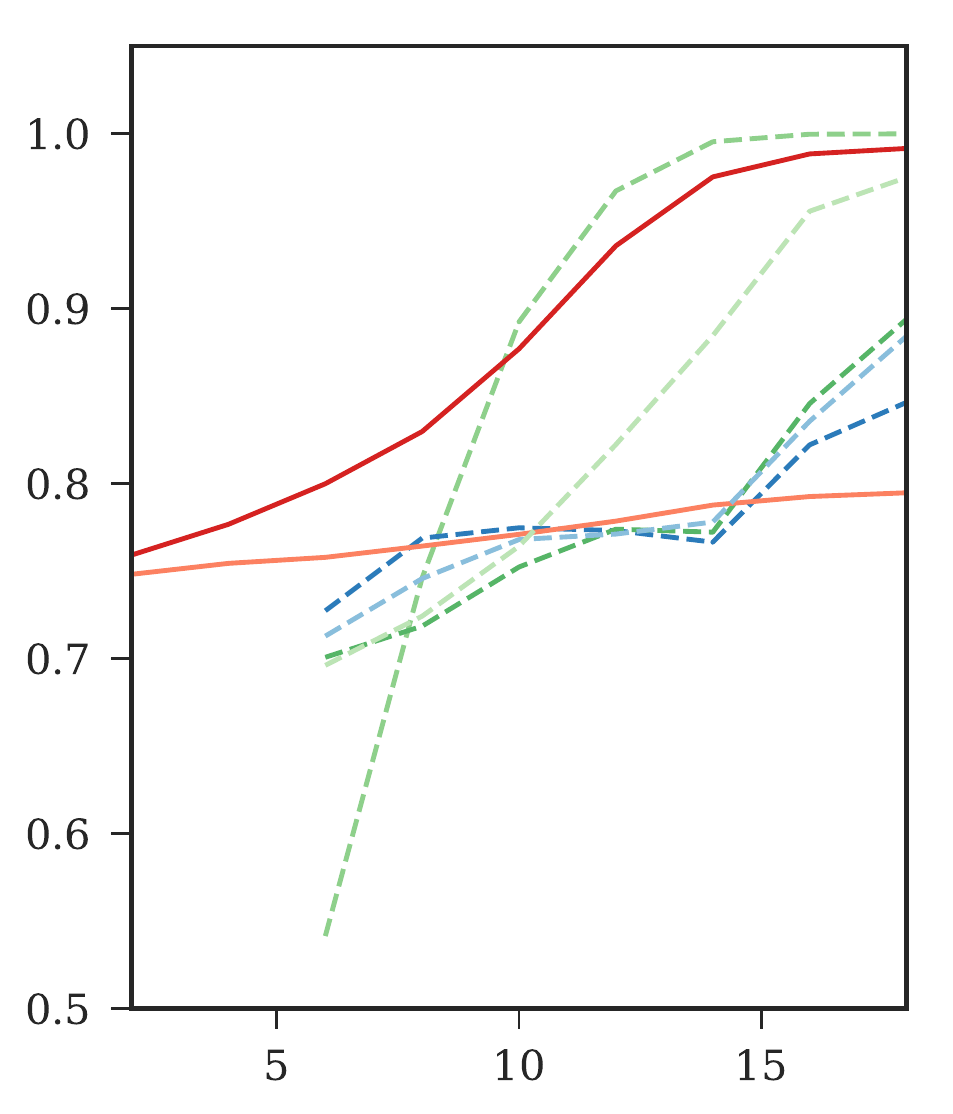}};
    \node[right=of Connect4train, node distance=0cm, xshift=-1.2cm]
          (Proteintrain)  {\includegraphics[width=0.24\linewidth]{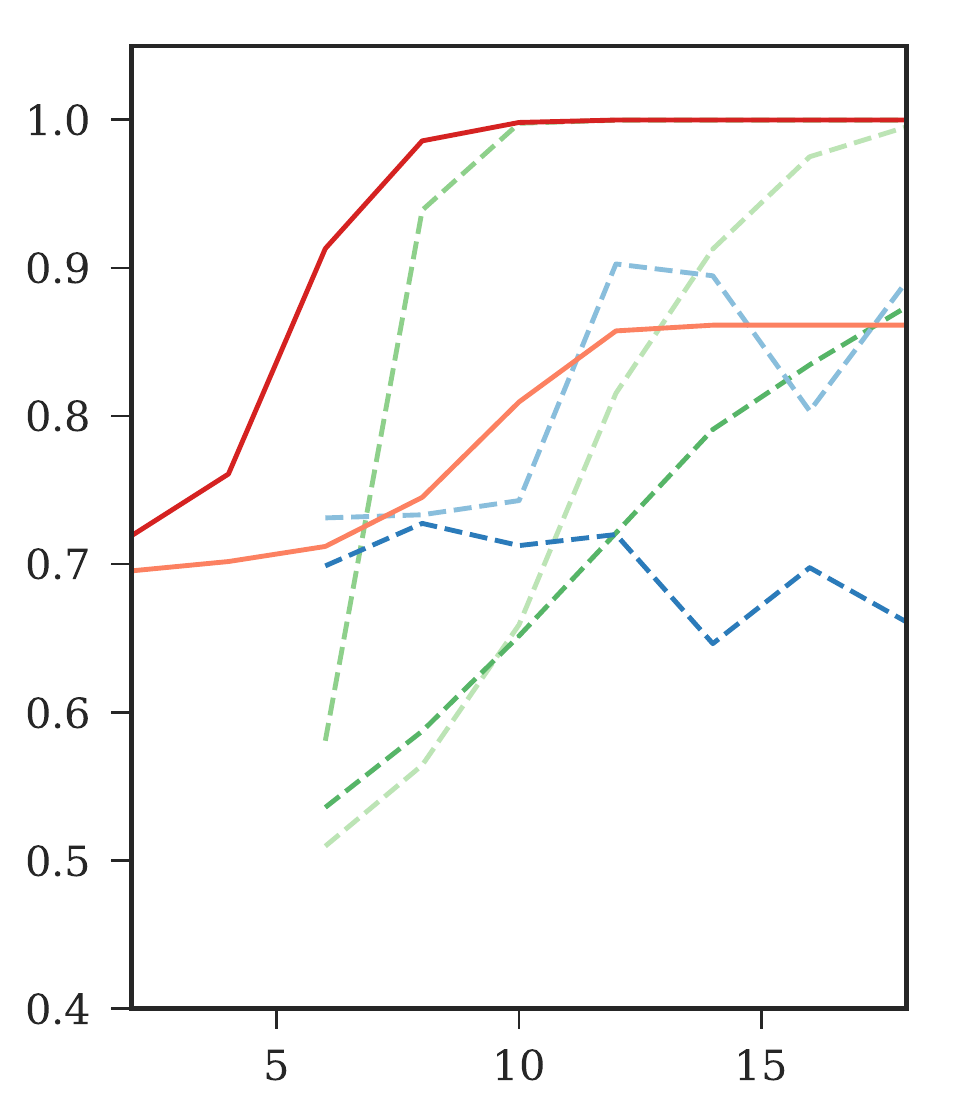}};

    \node[below=of MNISTtrain, node distance=0cm, font=\small,
          xshift=0.3cm, yshift=1.3cm] {Tree depth};
    \node[below=of SensITtrain, node distance=0cm, font=\small,
          xshift=0.3cm, yshift=1.3cm] {Tree depth};
    \node[below=of Connect4train, node distance=0cm, font=\small,
          xshift=0.3cm, yshift=1.3cm] {Tree depth};
    \node[below=of Proteintrain, node distance=0cm, font=\small,
          xshift=0.3cm, yshift=1.3cm] {Tree depth};

    \node[left=of MNISTtest, node distance=0cm, rotate=90, font=\small,
          anchor=center,yshift=-1.0cm] {Test accuracy};
    \node[left=of MNISTtrain, node distance=0cm, rotate=90, font=\small,
          anchor=center,yshift=-1.0cm] {Training accuracy};
  \end{tikzpicture}
  \begin{subfigure}[b]{0.24\textwidth}
    \caption{MNIST}
  \end{subfigure}
  \begin{subfigure}[b]{0.24\textwidth}
    \caption{SensIT}
  \end{subfigure}
  \begin{subfigure}[b]{0.24\textwidth}
    \caption{Connect4}
  \end{subfigure}
  \begin{subfigure}[b]{0.24\textwidth}
    \caption{Protein}
  \end{subfigure}
  \caption{Performance of deterministic oblique decision trees.
           Accuracy of \textit{Greedy} trained 
           (solid, light red line) and 
           \textit{Finetuned} oblique decision trees (solid, dark red line)
           on test and training sets is compared against other algorithms.
           The maximum tree depth varies from 2 to 18 with stepsize 2.
           Dashed lines represent results reported in \cite{Norouzi2015}.}
  \label{fig:results}
\end{figure*}
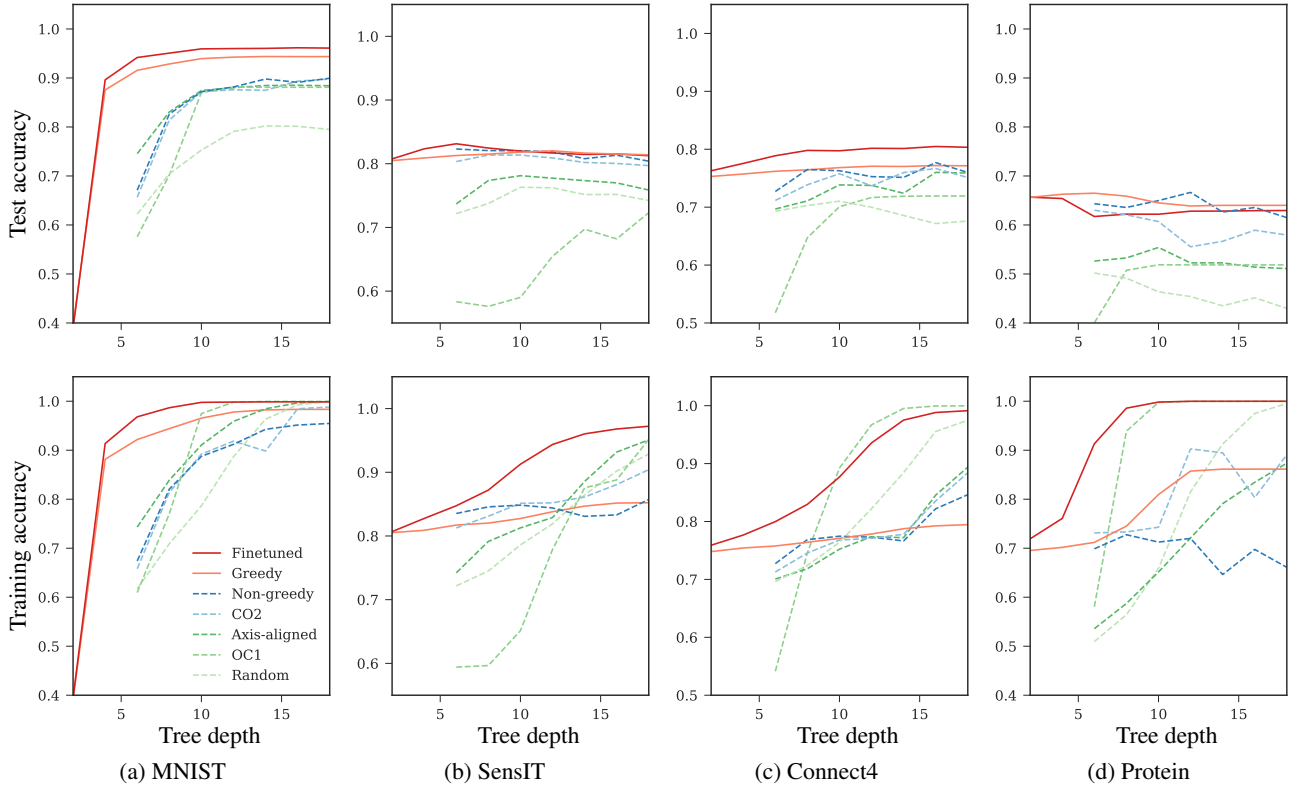

We conduct experiments on data from very different domains: first, on the
multivariate but unstructured datasets used in \cite{Norouzi2015} 
(section~\ref{sec:EXPaccuracy}).
Next, we show that the proposed algorithm can learn meaningful spatial 
features on \textit{MNIST}, \textit{FashionMNIST} and \textit{ISBI}, as has 
previously been demonstrated in neural networks but not in
decision trees (section~\ref{sec:EXPparamviz}).
Then, we demonstrate the same property on
a real-world biological image processing task (section~\ref{sec:EXPISBI}). 
Finally, we deliver proof of principle that a deterministic decision tree 
with complex split nodes can be trained end-to-end, by using a small 
neural network in each split node (section~\ref{sec:EXPCNN}). 

\subsection{Performance of oblique decision trees} \label{sec:EXPaccuracy}
We compare the performance of our algorithm in terms of accuracy to all results
reported in \cite{Norouzi2015}.
They only compare single, unpruned trees, since common 
ensemble methods such as bagging and boosting as well as pruning can be 
applied to all algorithms.
In order to provide a fair comparison, we also refrain from pruning,
ensembles and regularization.

\textbf{Datasets.}
Reference \cite{Norouzi2015} reports results on the following 
four datasets.
The multi-class classification datasets
\textit{SensIT (combined)}, 
\textit{Connect4}, \textit{Protein} and \textit{MNIST} are obtained from 
the LIBSVM repository \cite{LIBSVM}.
When a separate test set is not provided, we randomly split the data into
a training set with 80\% of the data and use 20\% for testing.
Likewise, when no validation set is provided, we randomly extract 20\%
of the training set as validation set.
In a preprocessing step, we normalize the data to zero mean and unit variance 
of the training data.

\textbf{Compared algorithms.}
The final model for prediction is always a deterministic decision tree with
either oblique or axis-aligned splits.
The following algorithms were evaluated in \cite{Norouzi2015}.
\textit{Axis-aligned}: conventional axis-aligned splits based on information
gain.
\textit{OC1}: oblique splits optimized with coordinate descent as proposed in
\cite{Murthy1996}.
\textit{Random}: selected the best of randomly generated oblique
splits based on information gain.
\textit{CO2}: greedy oblique tree algorithm based on structured
learning \cite{Norouzi2015CO2}.
\textit{Non-greedy}: non-greedy oblique decision tree algorithm based on
structured learning \cite{Norouzi2015}.
We compare the results of these algorithms with our proposed algorithms.
Here, \textit{Greedy} denotes a greedy initialization where each 
oblique split is computed using the EM optimization.
For each depth, we apply the \textit{Finetune} algorithm to the tree
obtained from the \textit{Greedy} algorithm at that depth.

\textbf{Hyperparameters and initialization.}
We keep all hyperparameters fixed and conduct a grid search only over the
number of training epochs in \(\{20, 35, 50, 65\}\), using a train/validation
split.
The test data is only used to report the final performance.
\\
For gradient-based split parameter optimization, we use the Adam optimizer \cite{Kingma2015}
with default parameters 
(\(\alpha = 0.001, \beta_1 = 0.9, \beta_2 = 0.999, \epsilon = 10^{-8}\)) and
a batch size of 1000 with shuffled batches.
The split steepness hyperparameter is set to \(\gamma = 1.0\) initially and
increased by 0.1 after each epoch 
(one epoch consists of the split parameter \(\bm{\theta}_{\bm{\beta}}\) 
updates of all training batches as well as the update of the leaf predictions
\(\bm{\theta}_{\bm{\pi}}\)).
\\
Initial split directions are sampled from the unit sphere and the
categorical leaf predictions are initialized uniformly random.

\textbf{Results.}
  Figure \ref{fig:results} shows the test and training statistical accuracy of
  the different decision tree learning algorithms.
  The accuracy of a classifier is defined as the ratio of correctly classified
  samples in the respective set.
  It was evaluated for a single tree at various maximum depths.
  The red solid lines show the result of our proposed algorithm, 
  the dashed lines represent results from~\cite{Norouzi2015}.
  \\
  Our algorithms achieve higher test accuracy 
  than previous work, especially in extremely shallow trees.
  The highest increase in test accuracy is observed on the \textit{MNIST} data
  set.
  Here, we significantly outperform previous approaches for oblique decision trees
  at all depths.
  In particular, an oblique decision tree of depth 4 is already sufficient to
  surpass all competitors.
  \\
  Likewise, on \textit{Connect4} our approach performs better at all depths.
  Notably, a decision tree of depth 2 is sufficient to exceed previous
  approaches.
  \\
  On \textit{SensIT} and \textit{Protein} we perform better than or on par with the
  \textit{Non-greedy} approach proposed in \cite{Norouzi2015}.
  However, experiments with regularization of leaf features have shown that with more
  hyperparameter tuning overfitting may be reduced, \eg on the \textit{Protein}
  dataset and thus the results may be improved.
  We did not include this here, as we aimed to provide a fair comparison and 
  show the performance given very little fine-tuning.
  \\
  Generally, our algorithm also trains more accurate oblique decision trees
  on of the depth complexity on the training data.
  
  In conclusion, our experiments show that our proposed algorithm is able to learn
  more accurate deterministic oblique decision trees than previous approaches.
  Furthermore, we refrained from hyperparameter tuning to show that
  the approach even works well with default parameters.
\begin{figure*}[t]
  \centering
  \textit{MNIST}~~~~~~~~~~~~~
  \begin{subfigure}{0.13\textwidth}
    \includegraphics[width=\textwidth]{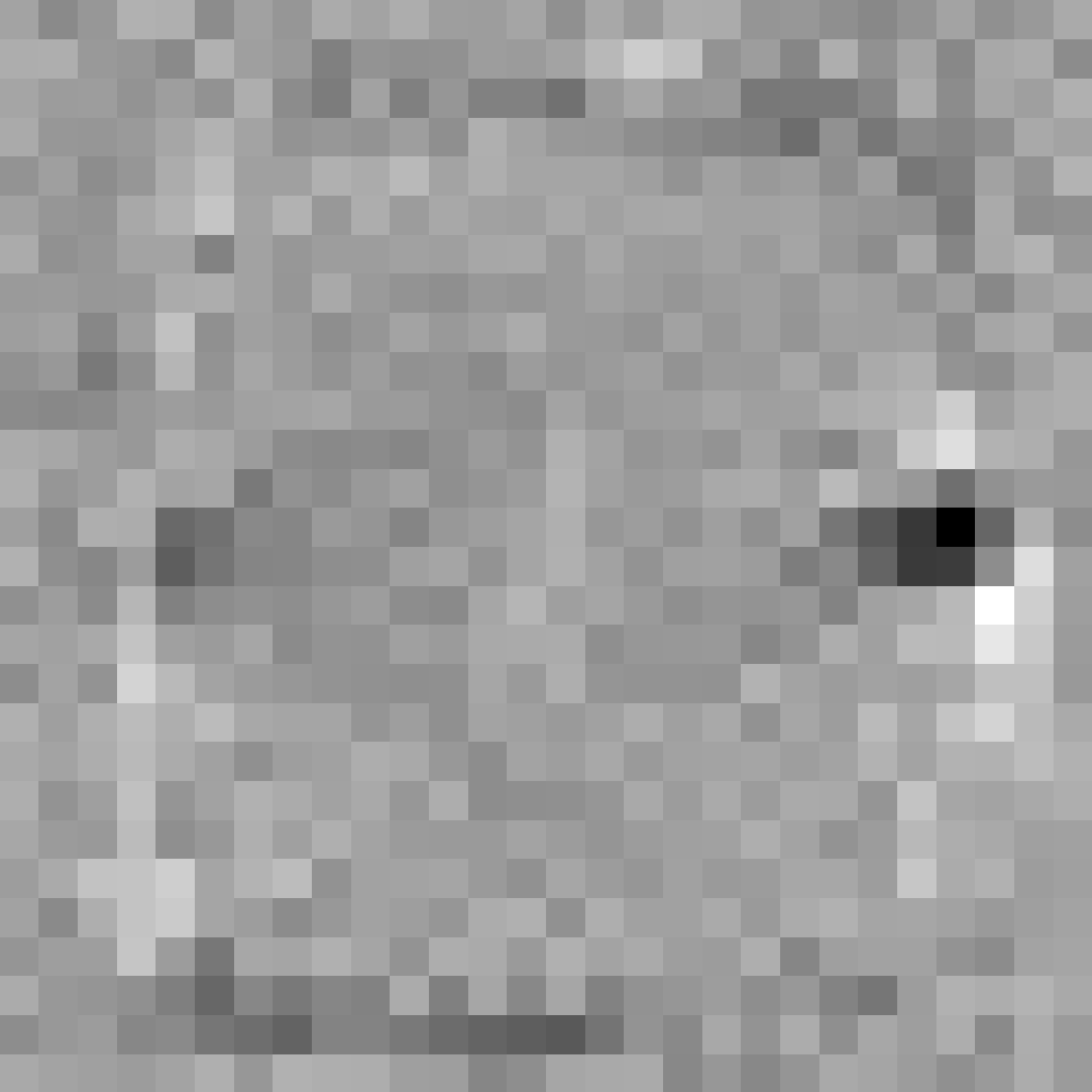}
  \end{subfigure}
  \begin{subfigure}{0.13\textwidth}
    \includegraphics[width=\textwidth]{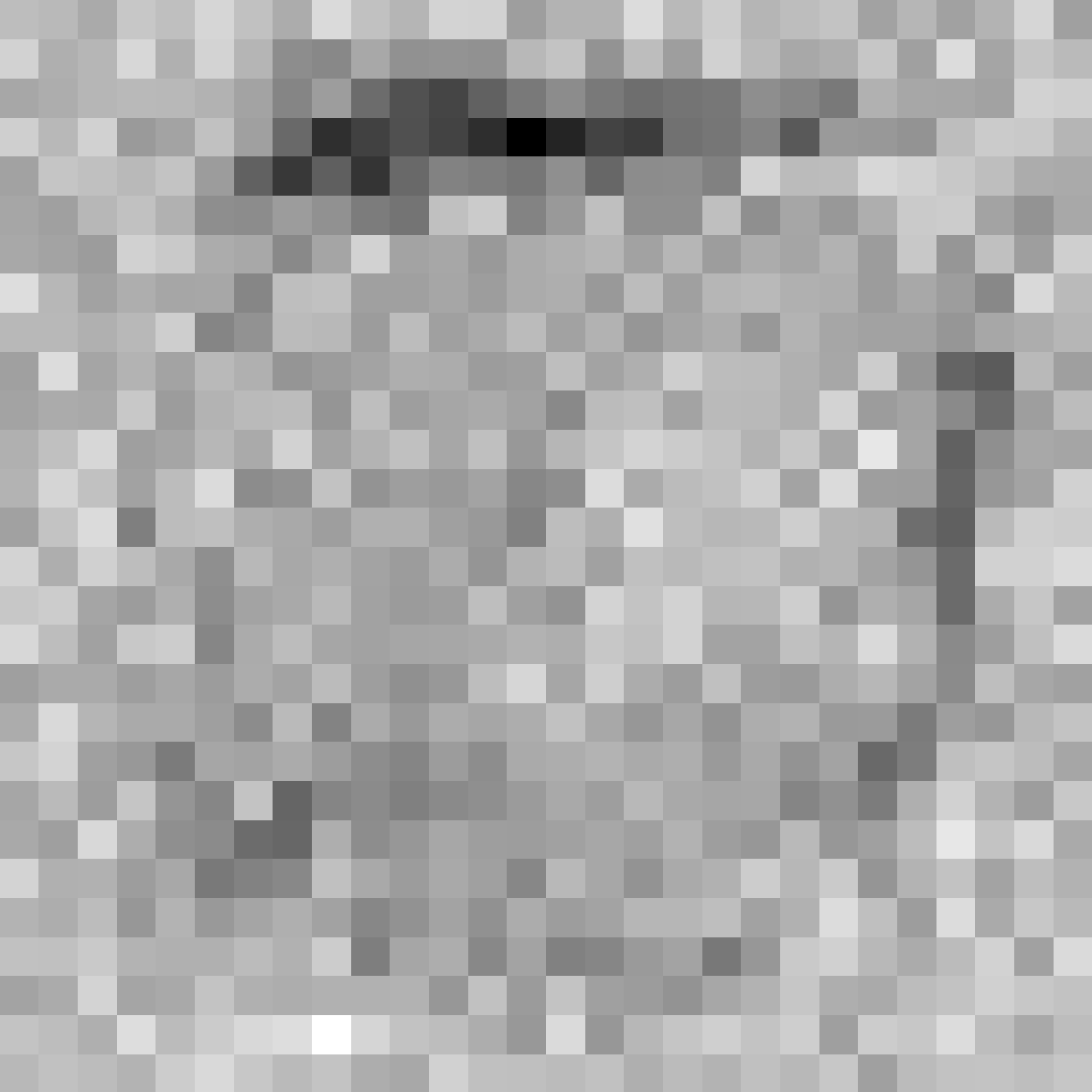}
  \end{subfigure}
  ~
  ~
  \begin{subfigure}{0.13\textwidth}
    \includegraphics[width=\textwidth]{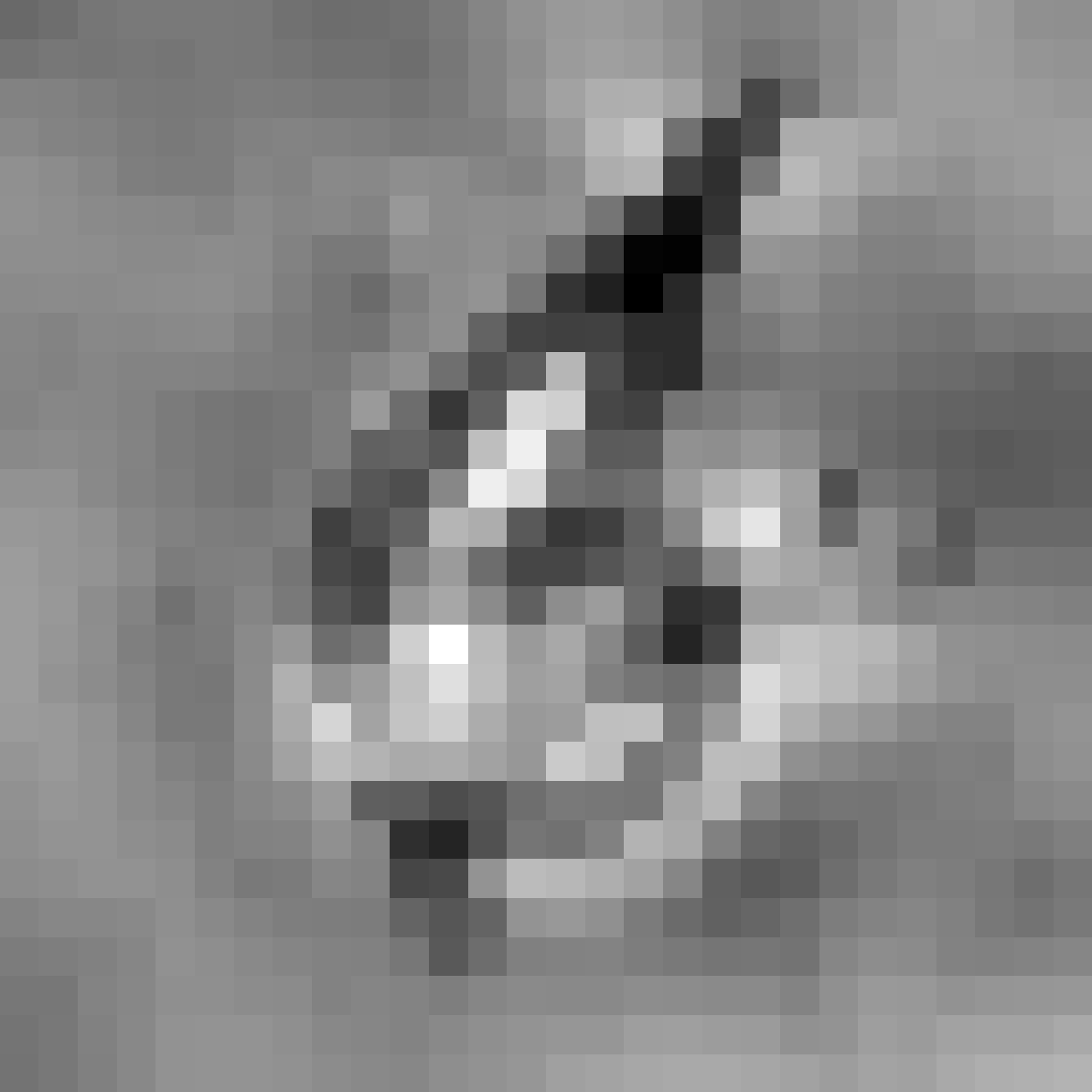}
  \end{subfigure}
  \begin{subfigure}{0.13\textwidth}
    \includegraphics[width=\textwidth]{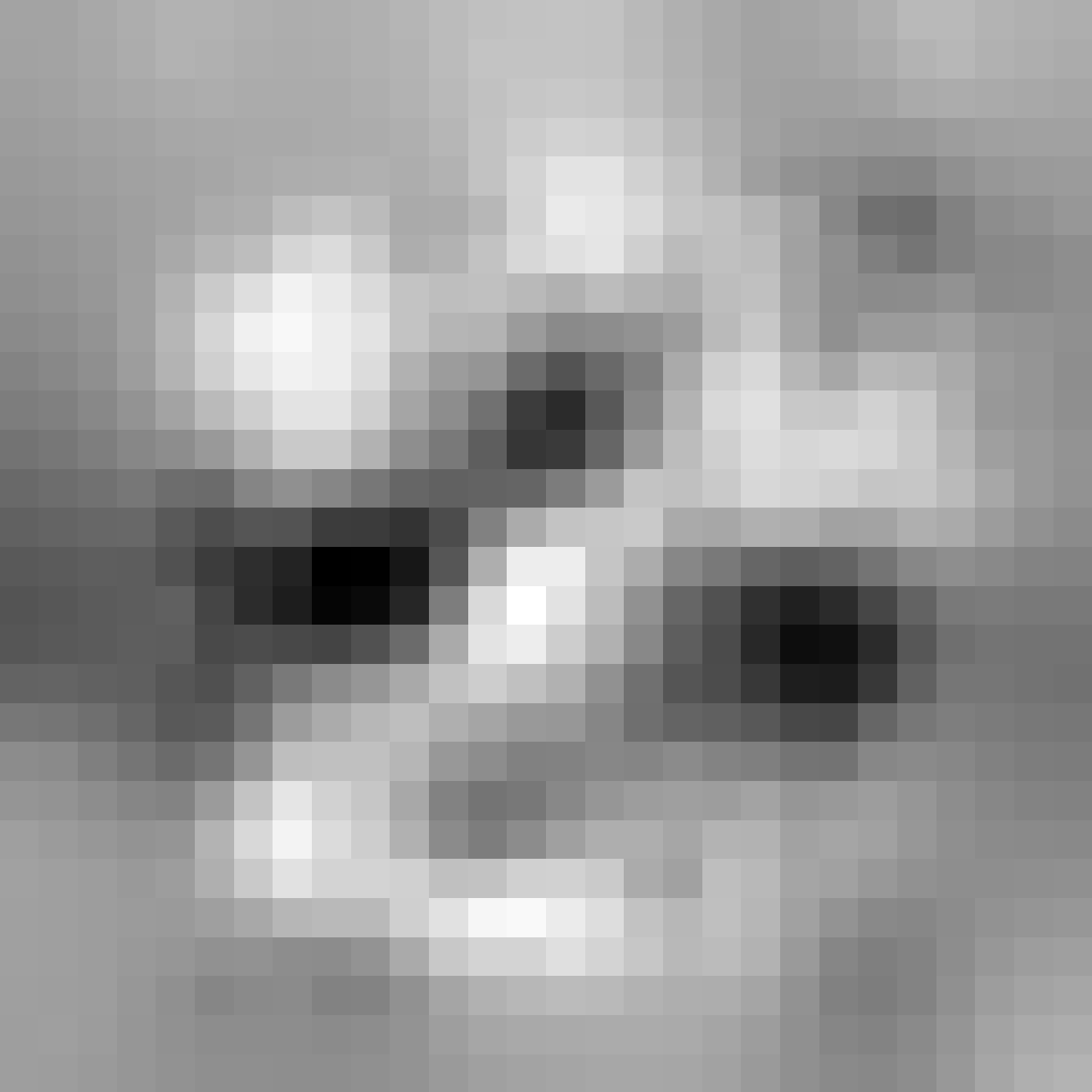}
  \end{subfigure}
  \begin{subfigure}{0.13\textwidth}
    \includegraphics[width=\textwidth]{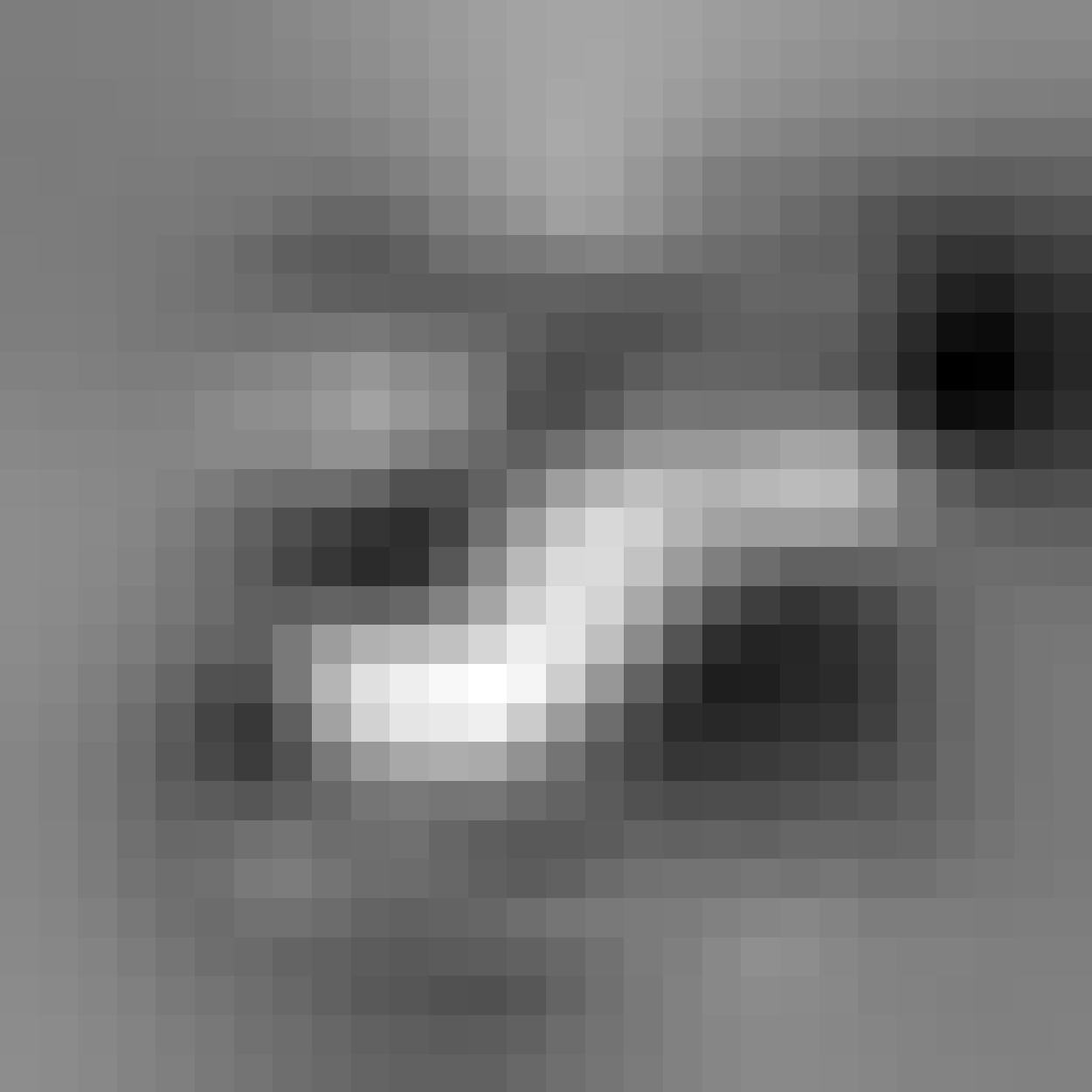}
  \end{subfigure}
  \begin{subfigure}{0.13\textwidth}
    \includegraphics[width=\textwidth]{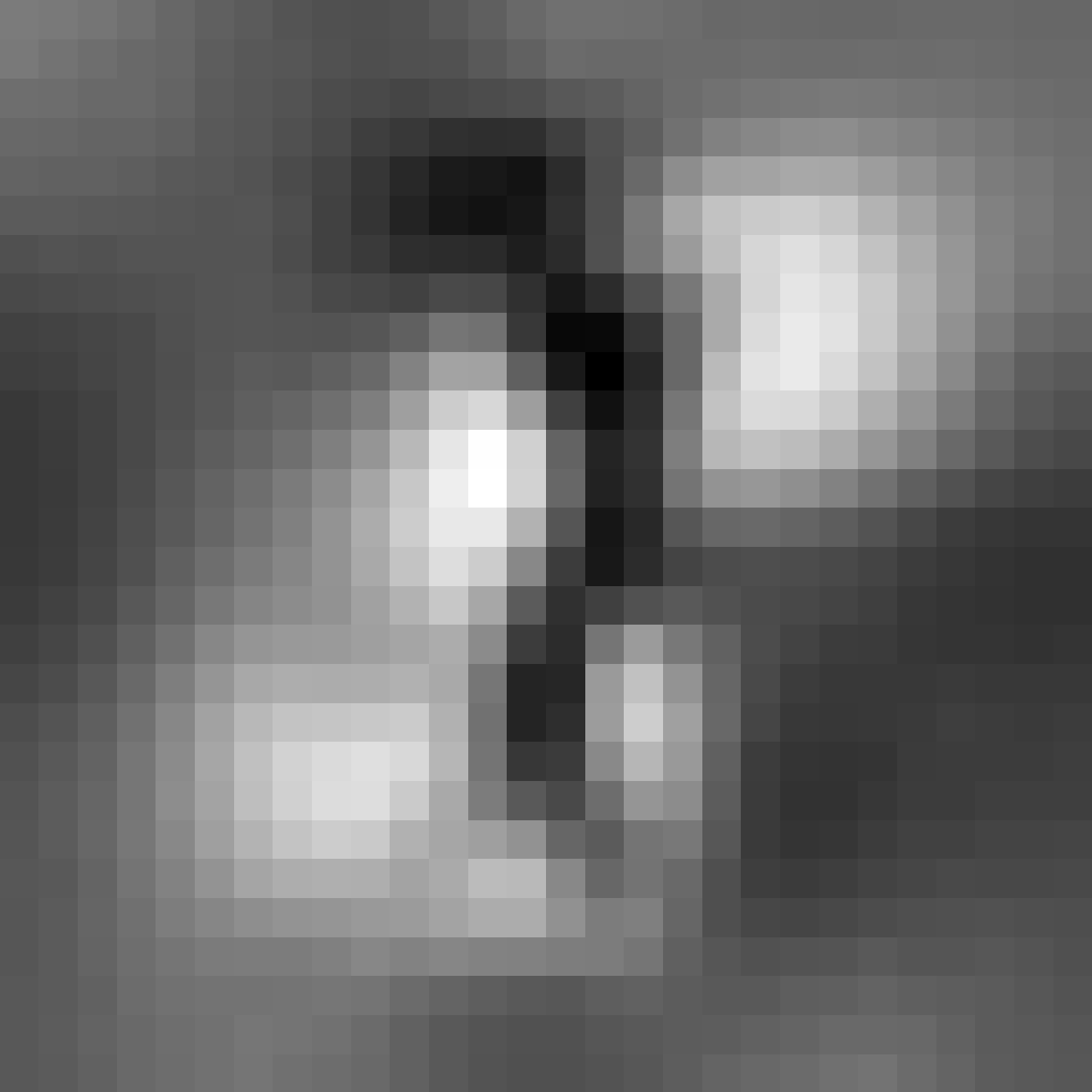}
  \end{subfigure}
  \\
  \textit{FashionMNIST}
  \begin{subfigure}{0.13\textwidth}
    \includegraphics[width=\textwidth]{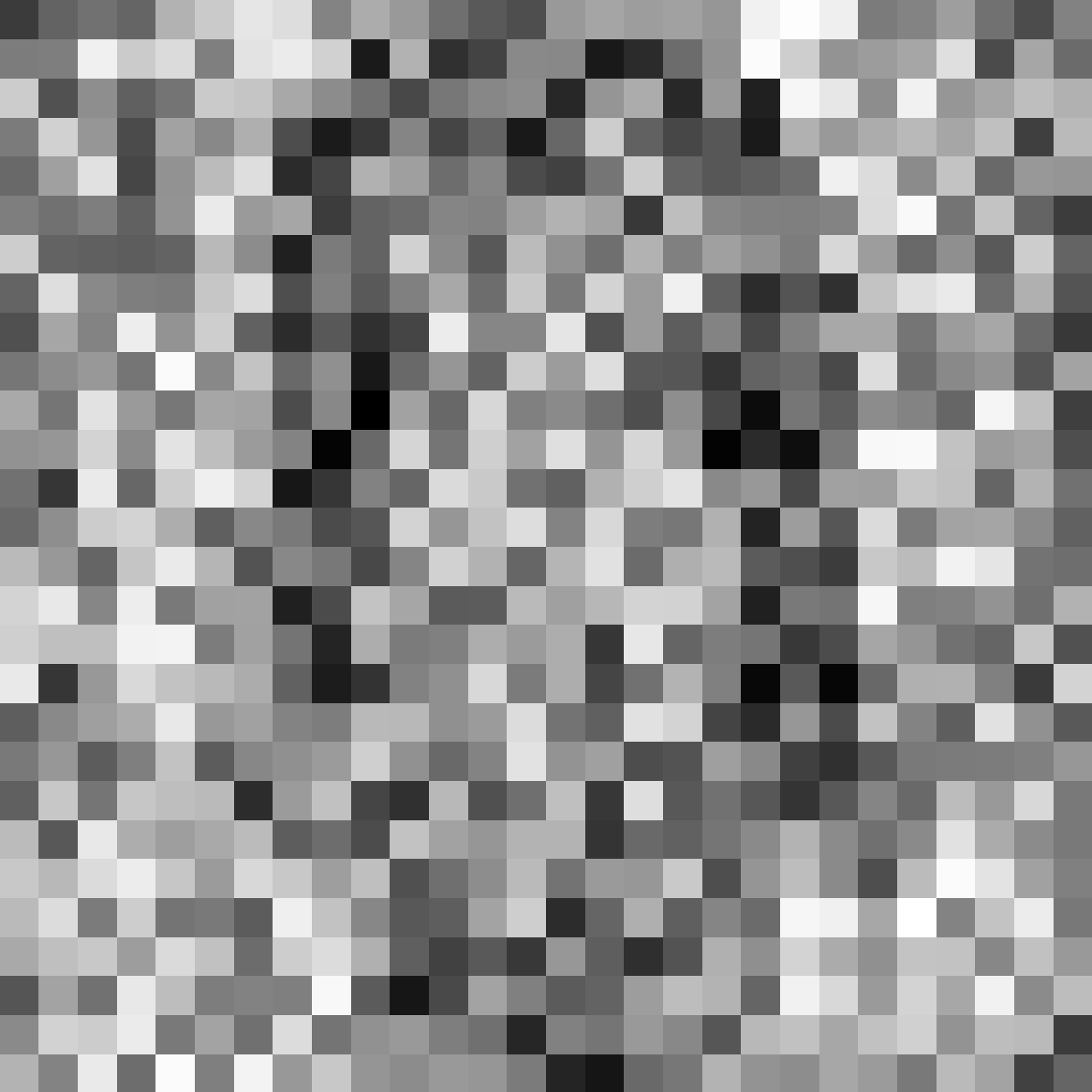}
  \end{subfigure}
  \begin{subfigure}{0.13\textwidth}
    \includegraphics[width=\textwidth]{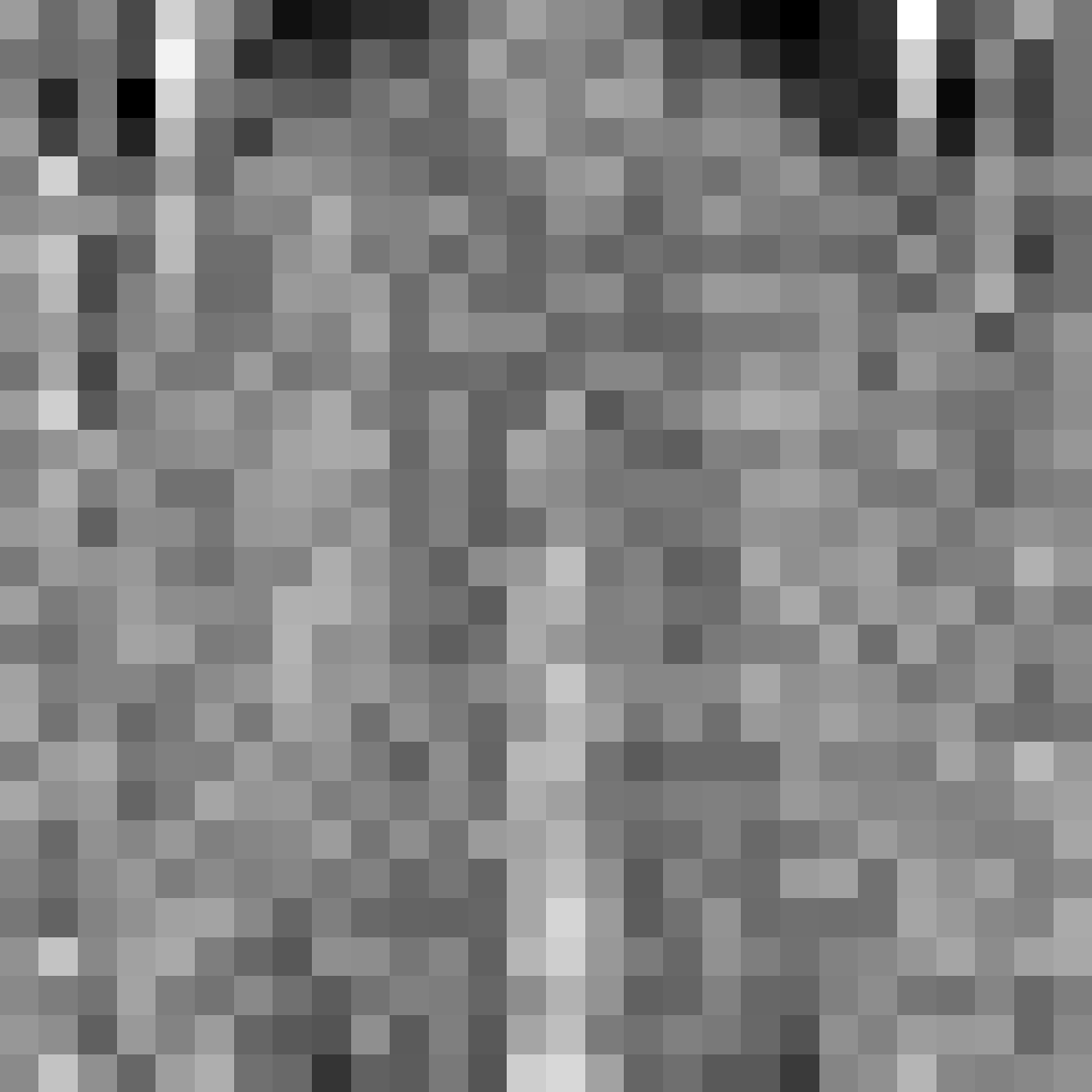}
  \end{subfigure}
  ~
  ~
  \begin{subfigure}{0.13\textwidth}
    \includegraphics[width=\textwidth]{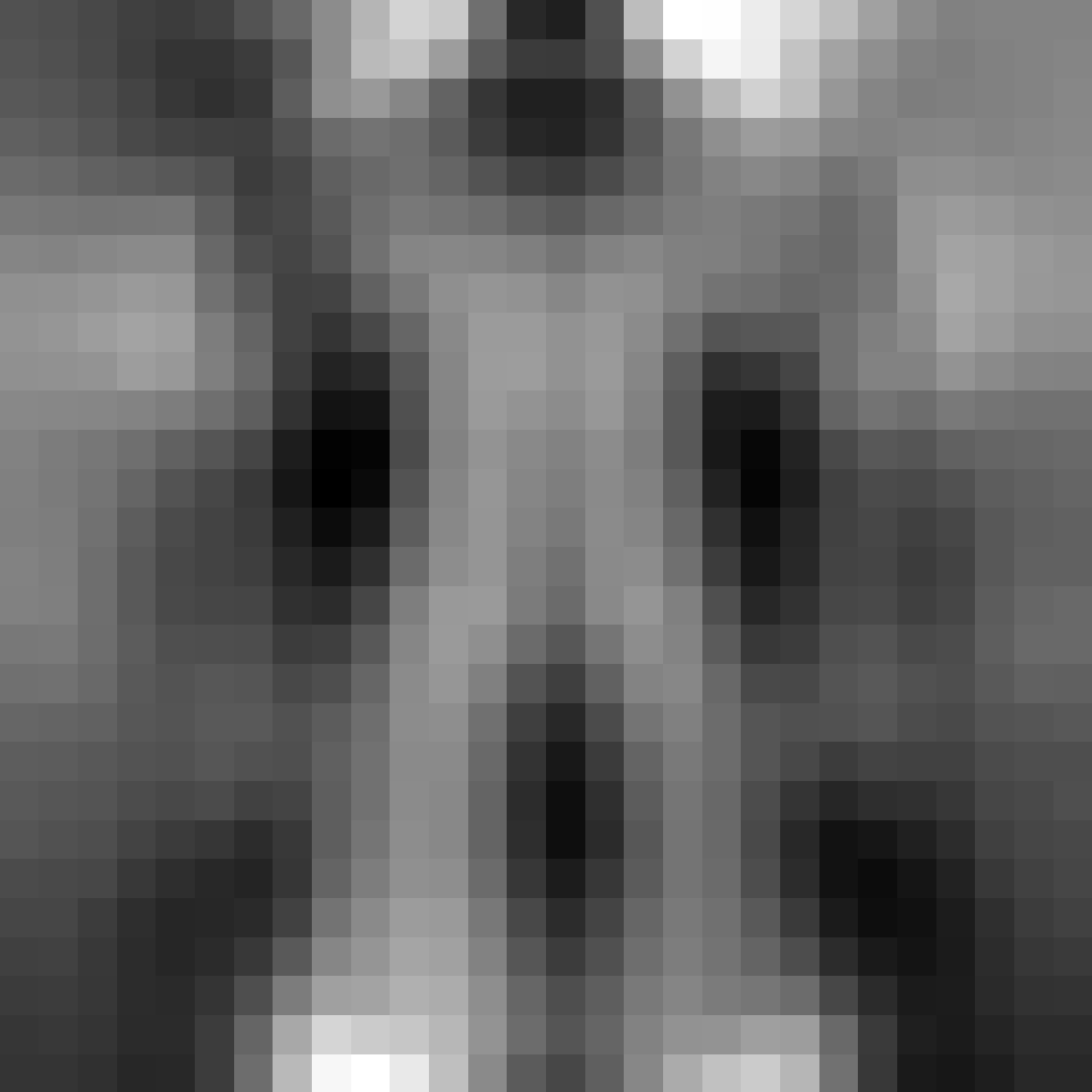}
  \end{subfigure}
  \begin{subfigure}{0.13\textwidth}
    \includegraphics[width=\textwidth]{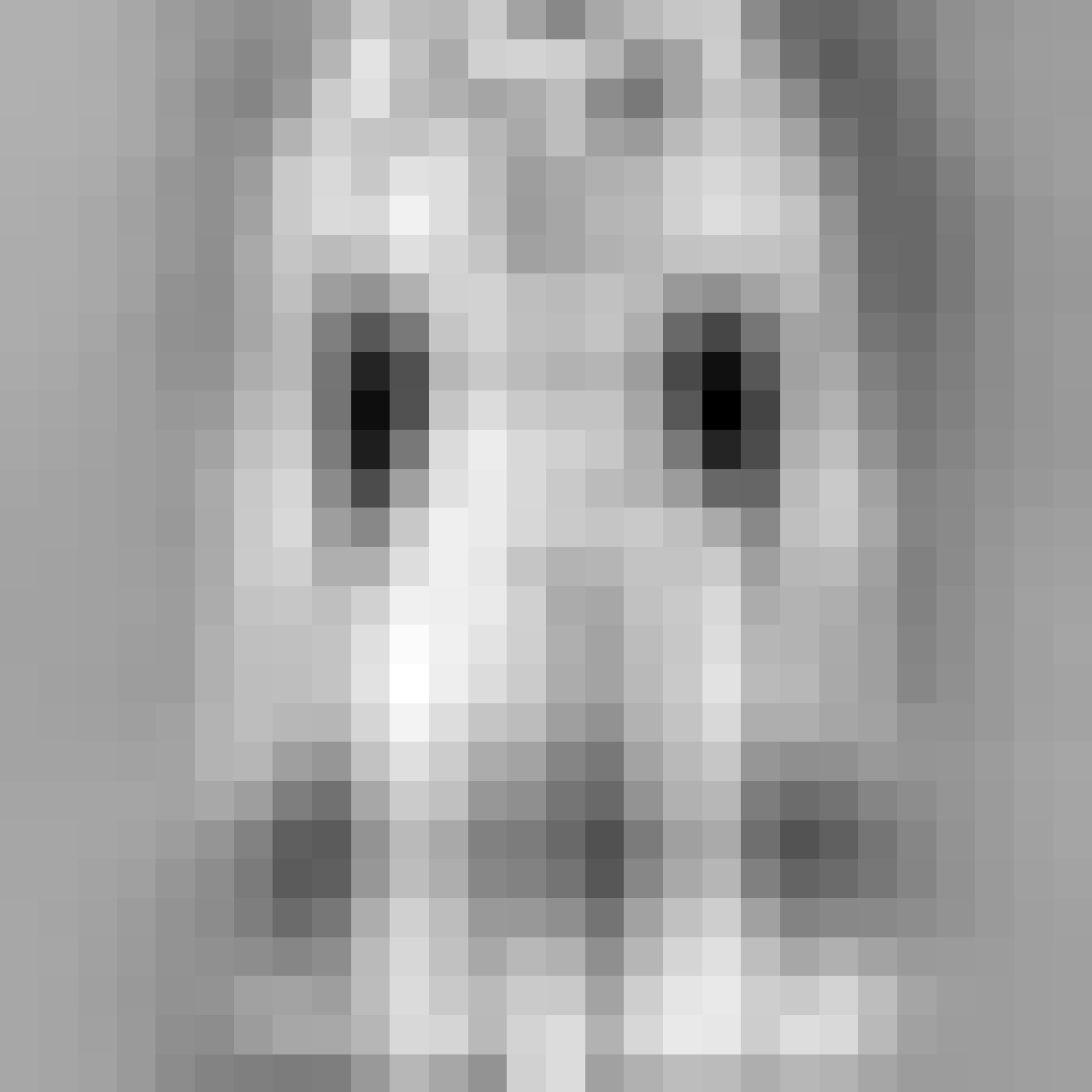}
  \end{subfigure}
  \begin{subfigure}{0.13\textwidth}
    \includegraphics[width=\textwidth]{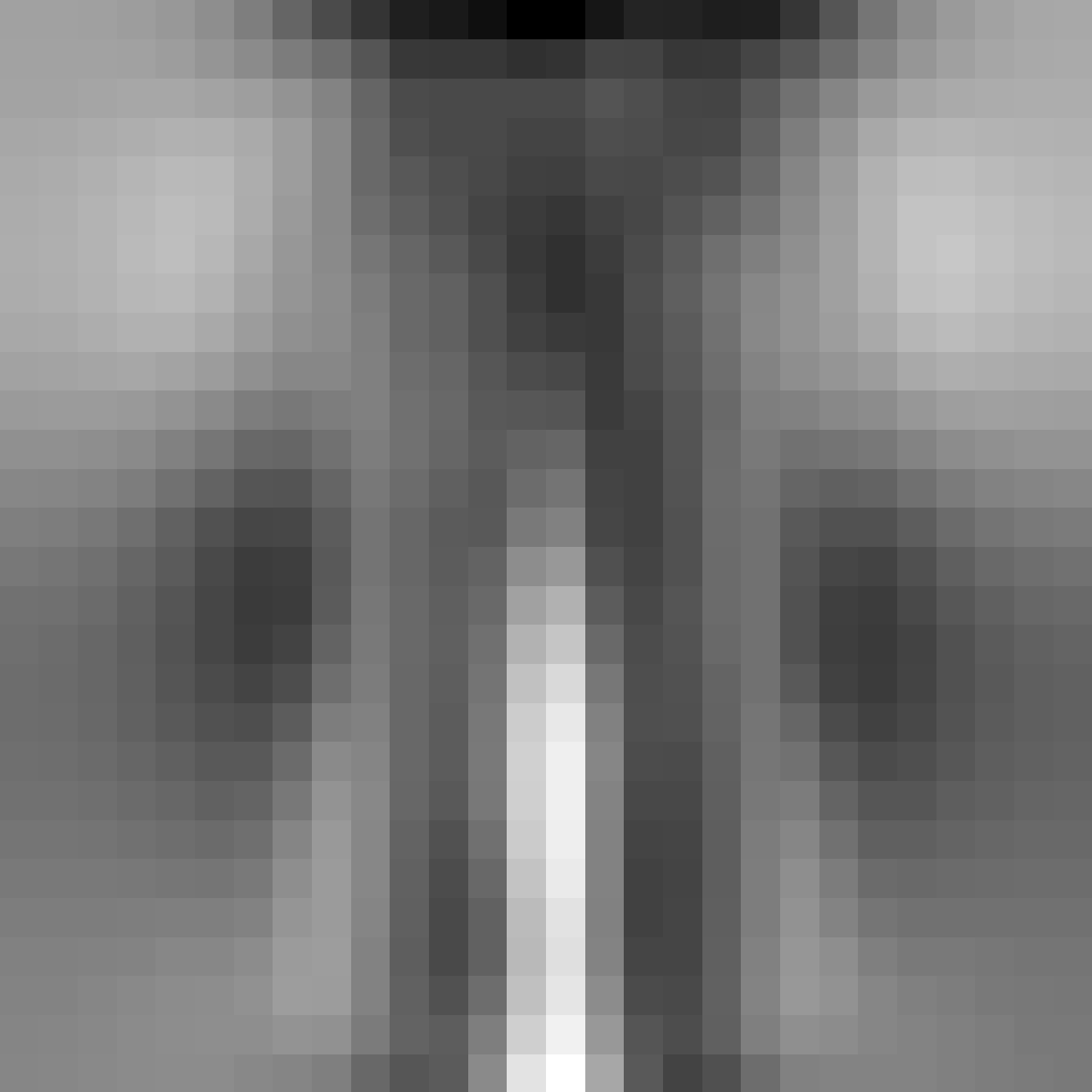}
  \end{subfigure}
  \begin{subfigure}{0.13\textwidth}
    \includegraphics[width=\textwidth]{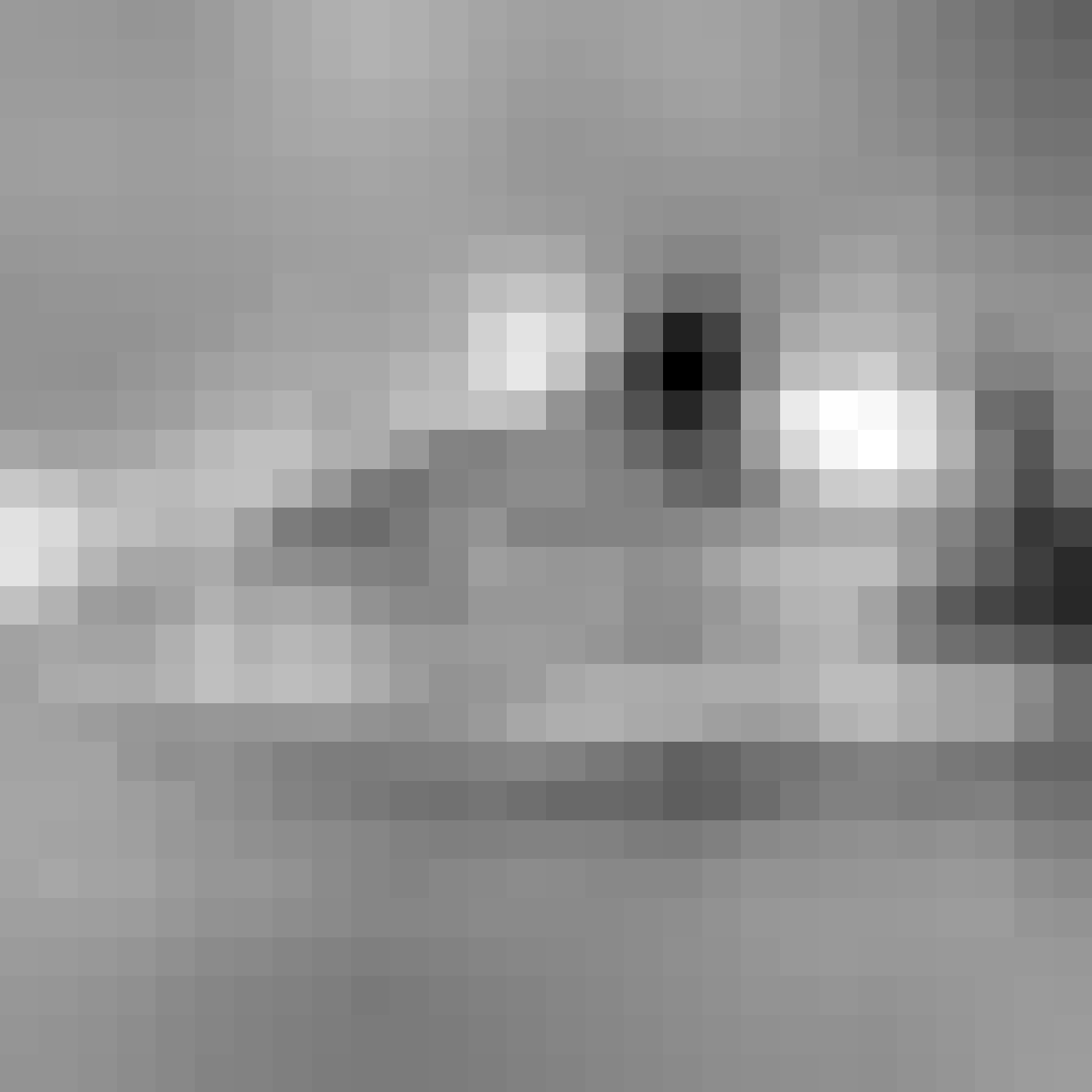}
  \end{subfigure}
  \\
  \textit{ISBI}~~~~~~~~~~~~~~~~~
  \begin{subfigure}{0.13\textwidth}
    \includegraphics[width=\textwidth]{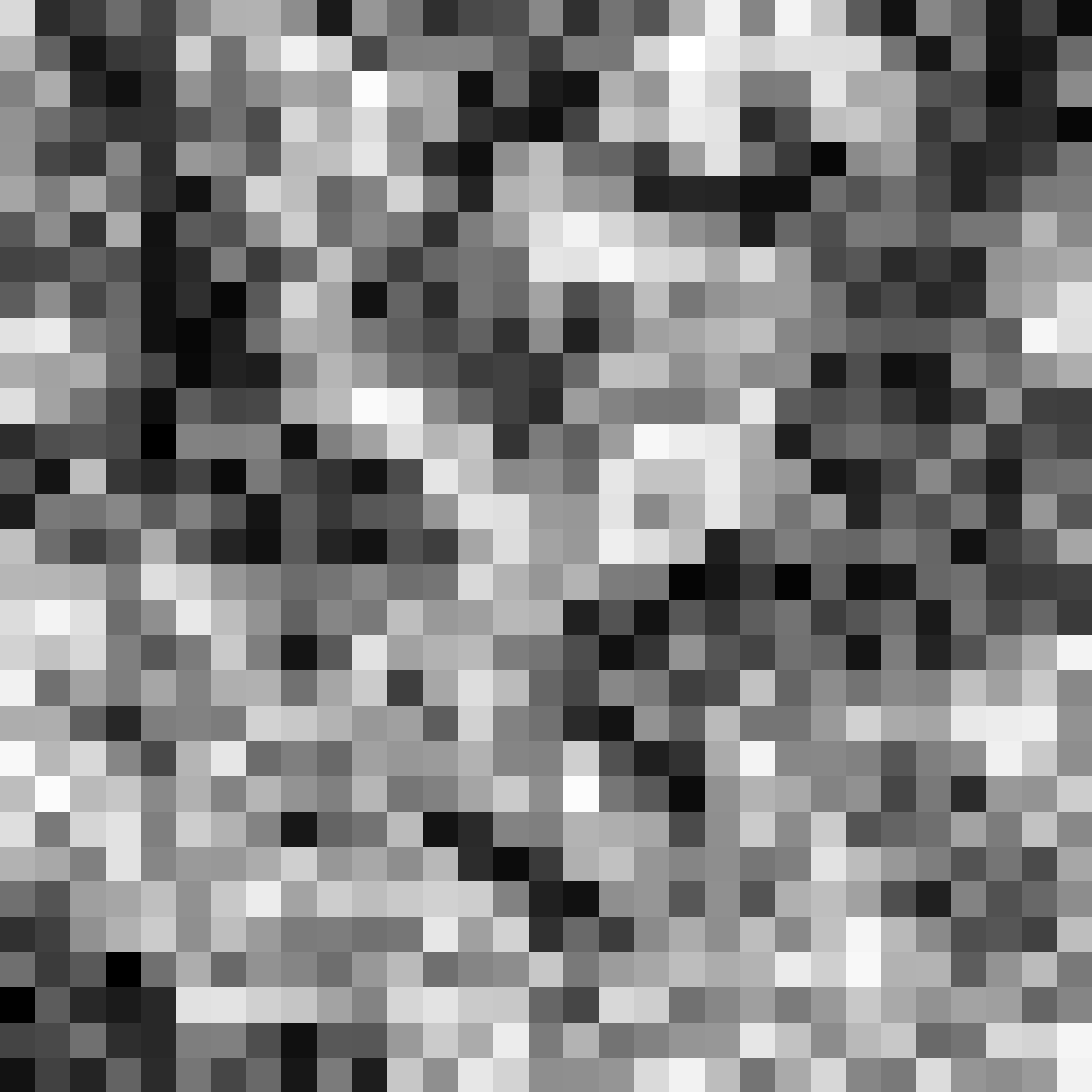}
  \end{subfigure}
  \begin{subfigure}{0.13\textwidth}
    \includegraphics[width=\textwidth]{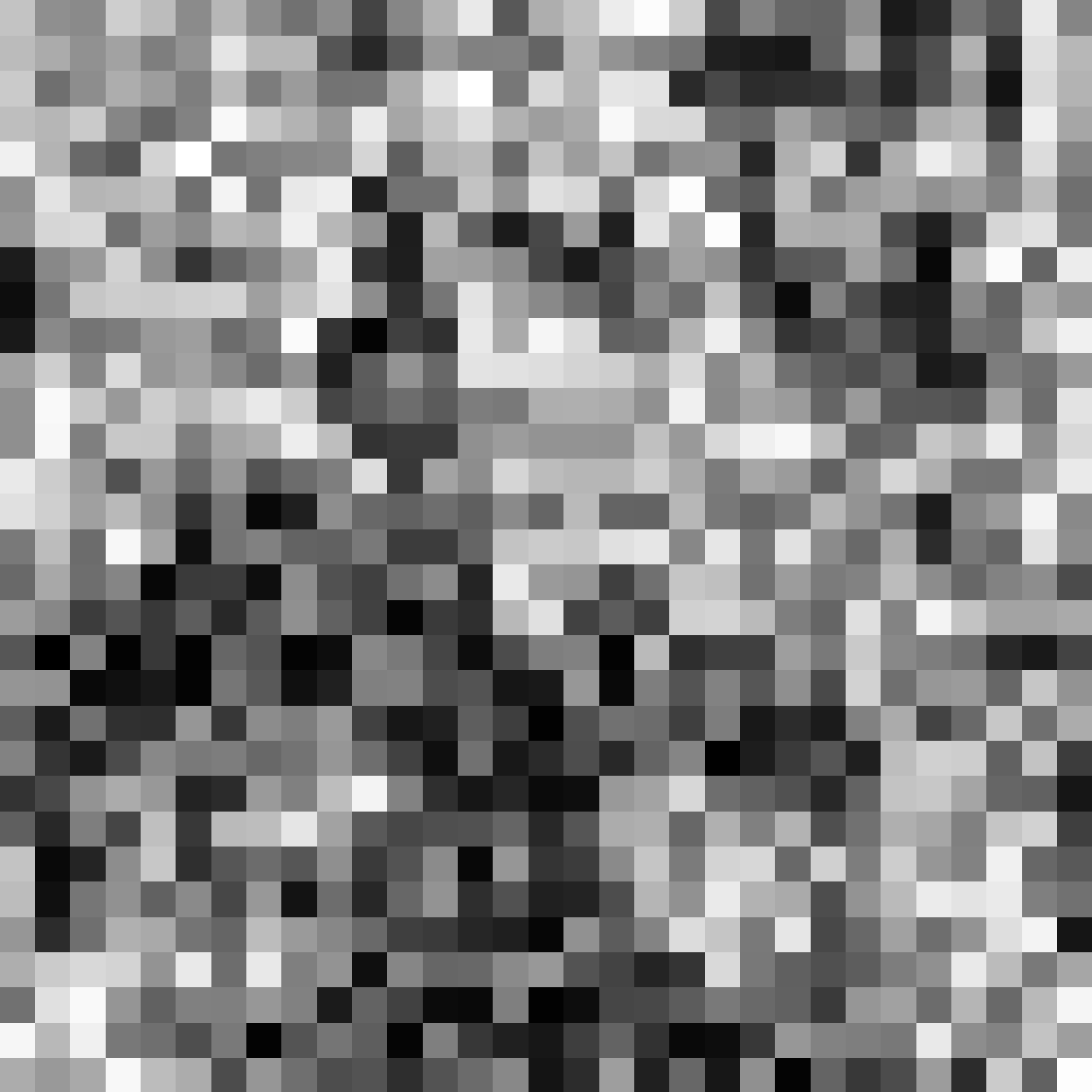}
  \end{subfigure}
  ~
  ~
  \begin{subfigure}{0.13\textwidth}
    \includegraphics[width=\textwidth]{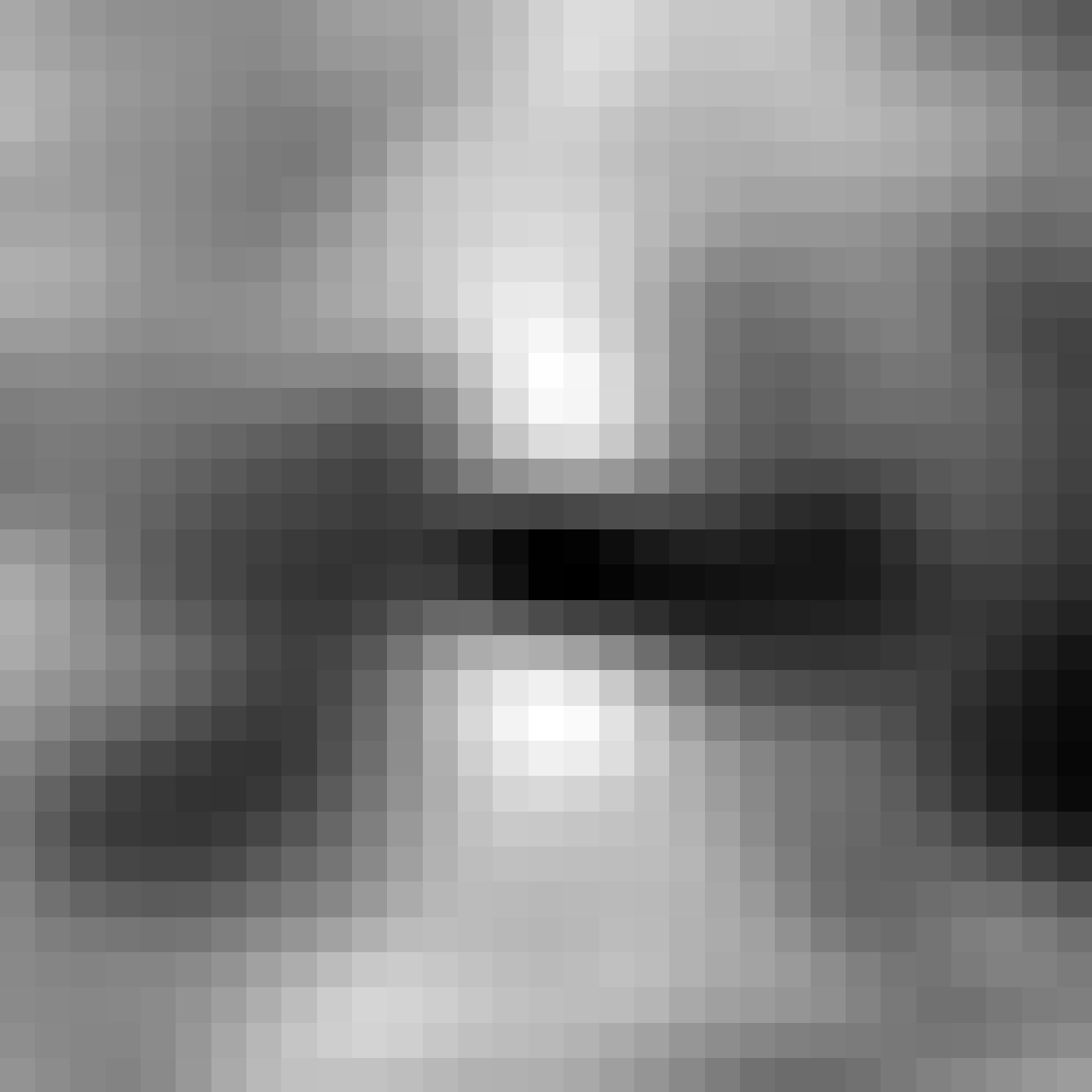}
  \end{subfigure}
  \begin{subfigure}{0.13\textwidth}
    \includegraphics[width=\textwidth]{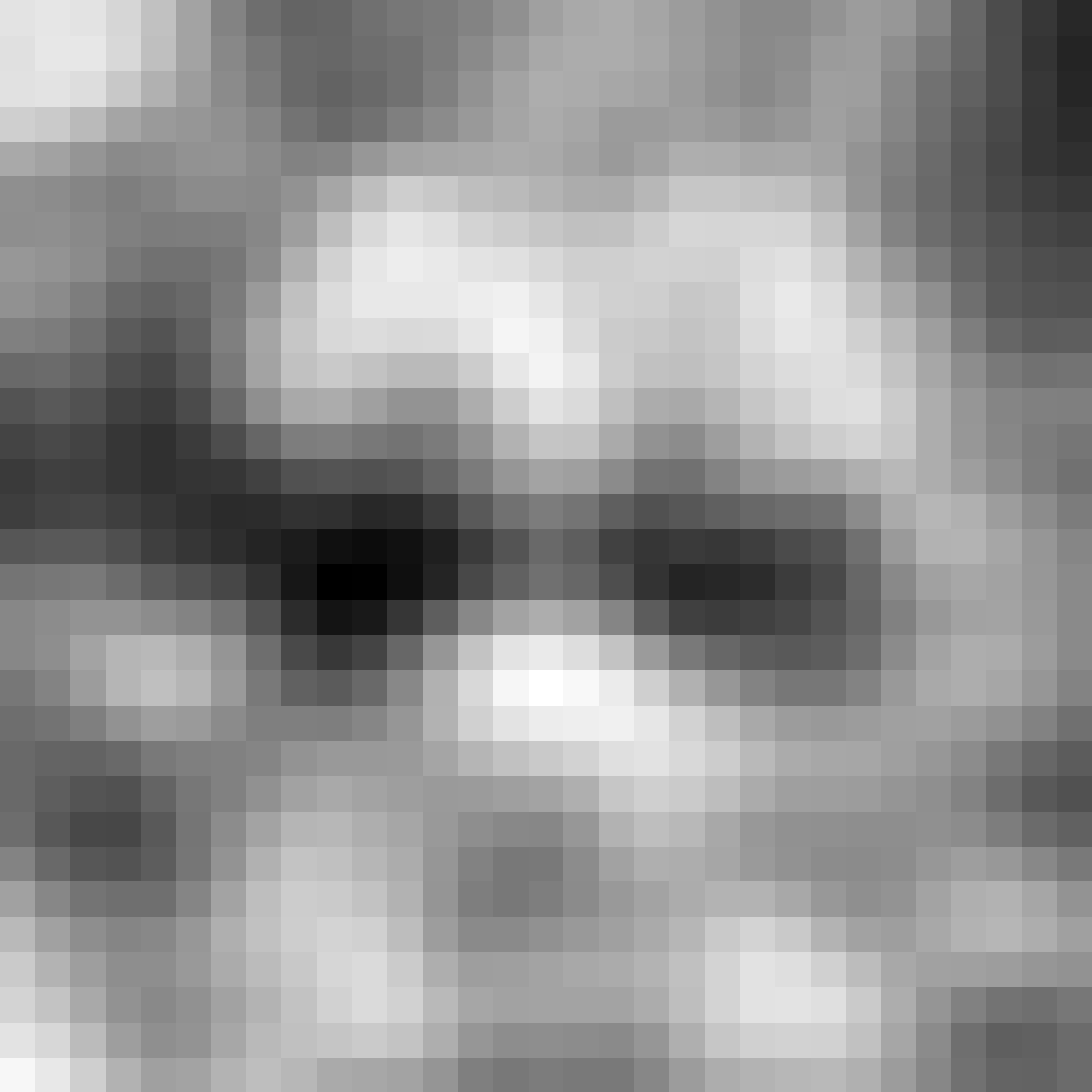}
  \end{subfigure}
  \begin{subfigure}{0.13\textwidth}
    \includegraphics[width=\textwidth]{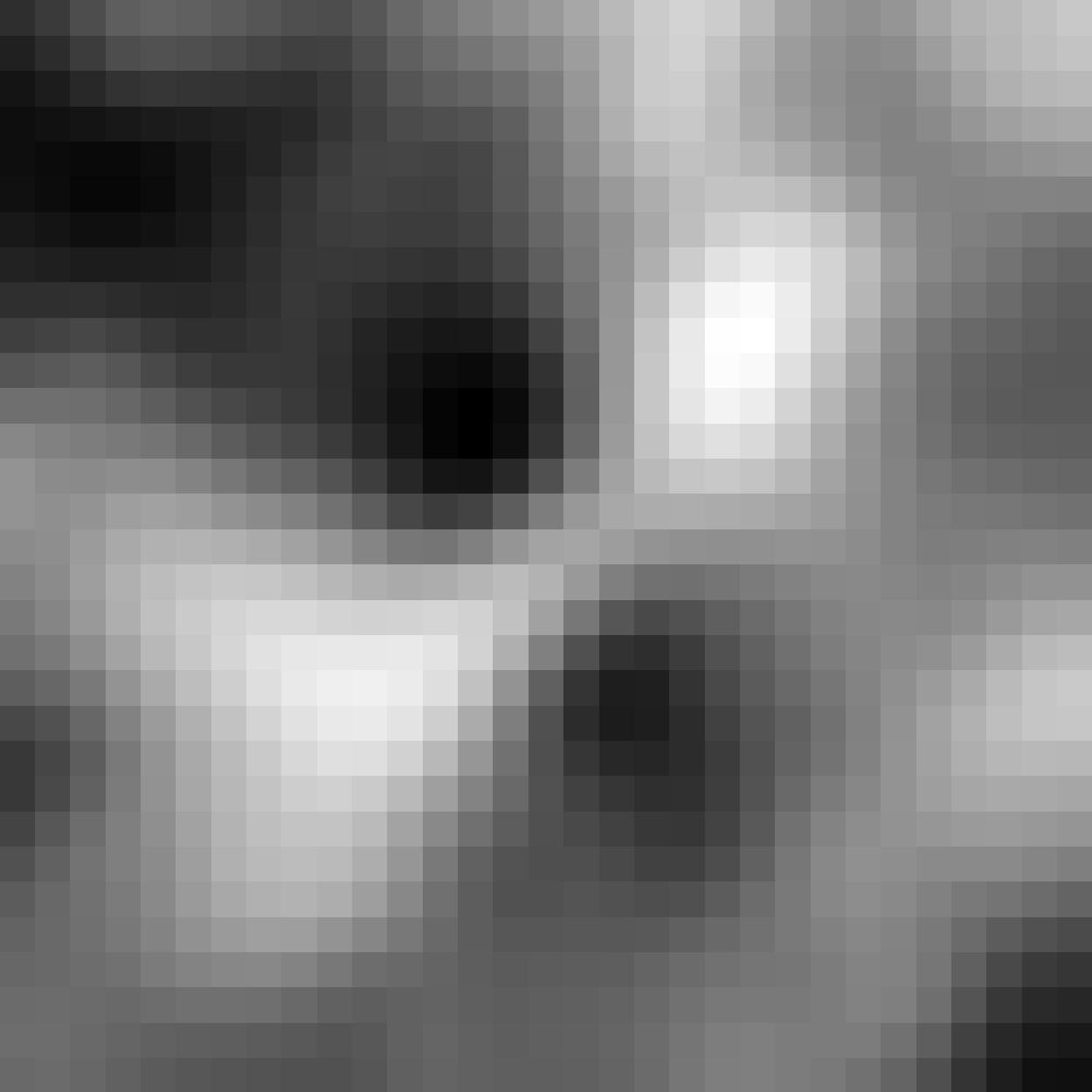}
  \end{subfigure}
  \begin{subfigure}{0.13\textwidth}
    \includegraphics[width=\textwidth]{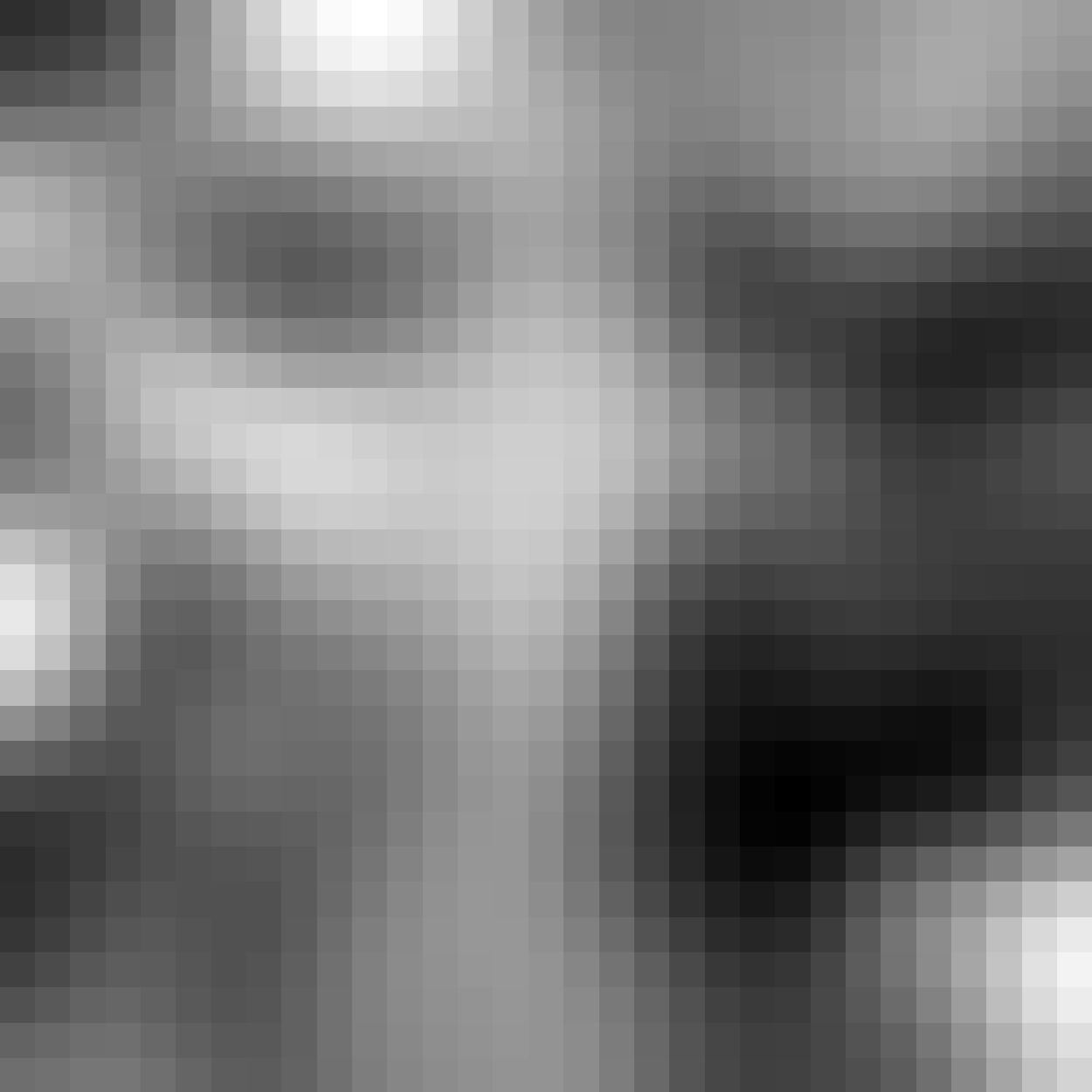}
  \end{subfigure}
  \\
  \hfill
  \begin{subfigure}{0.24\textwidth}
    \caption{Without spatial regularization}
  \end{subfigure}
  \begin{subfigure}{0.59\textwidth}
    \caption{With spatial regularization}
  \end{subfigure}
  \caption{Visualizations of oblique split parameters learned with and 
           without spatial regularization (section~\ref{sec:METHsplits}).
           The parameters are learned on different datasets, viz.\
           \textit{MNIST}~\cite{LeCun1998} (top row), 
           \textit{FashionMNIST}~\cite{FashionMNIST} (center row) and
           \textit{ISBI}~\cite{ISBI} (bottom row).
           Parameters trained with spatial regularization show visible
           structures and patterns, whereas parameters learned without
           regularization appear noisy.
           In both cases, we selected the parameters that show the best 
           visible structures.}
  \label{fig:MNISTfiltersshallow}
\end{figure*}

\subsection{Spatially regularized parameters} \label{sec:EXPparamviz}
We now investigate the effects of spatial regularization
(section~\ref{sec:METHsplits}) on the parameters of oblique
decision trees learned with our algorithm.
For this purpose, we train oblique decision trees on the \textit{MNIST}
digit dataset~\cite{LeCun1998}, the \textit{FashionMNIST} fashion product
dataset~\cite{FashionMNIST} and the \textit{ISBI} image partitioning
dataset~\cite{ISBI} comprising serial section Transmission Electron Microscopy
images.
In figure~\ref{fig:MNISTfiltersshallow}, we visualized selected parameters of
the oblique splits at various depths with and without regularization.
In both cases, we selected the parameters that displayed the best visible
structures.
For \textit{MNIST} and \textit{FashionMNIST}, the parameters were reshaped to
\(28\times28\) images, such that each parameter pixel corresponds to the
respective pixel in the training images.
To solve the segmentation task on \textit{ISBI}, we provide a sliding window
of size \(31 \times 31\) as features for each pixel in the center of the
window.
Moreover, we linearly normalized the parameters to the full grayscale range.

\textbf{Results.} Regularization penalizes differences in adjacent parameters.
The parameters without regularization appear very noisy and
it is difficult for the human eye to identify structures.
In contrast, in the experiments with
regularization the algorithm learns smoother parameter patterns, without
decreasing the accuracy of the decision trees.
The regularized parameters display structures and recognizable patterns.
The patterns learned on the \textit{MNIST} show visible
sigmoidal shapes and even recognizable digits.
On the \textit{FashionMNIST} dataset, the regularized parameters
display the silhouettes of coats, pants and sneakers.
Likewise, our algorithm is able to learn the structures of membranes on
the real-world biological electron microscopy images from the \textit{ISBI}
dataset.

\subsection{Image segmentation} \label{sec:EXPISBI}
\begin{figure*}[ht!] 
  \centering
  \begin{subfigure}[b]{0.24\textwidth}
    \includegraphics[width=\linewidth]{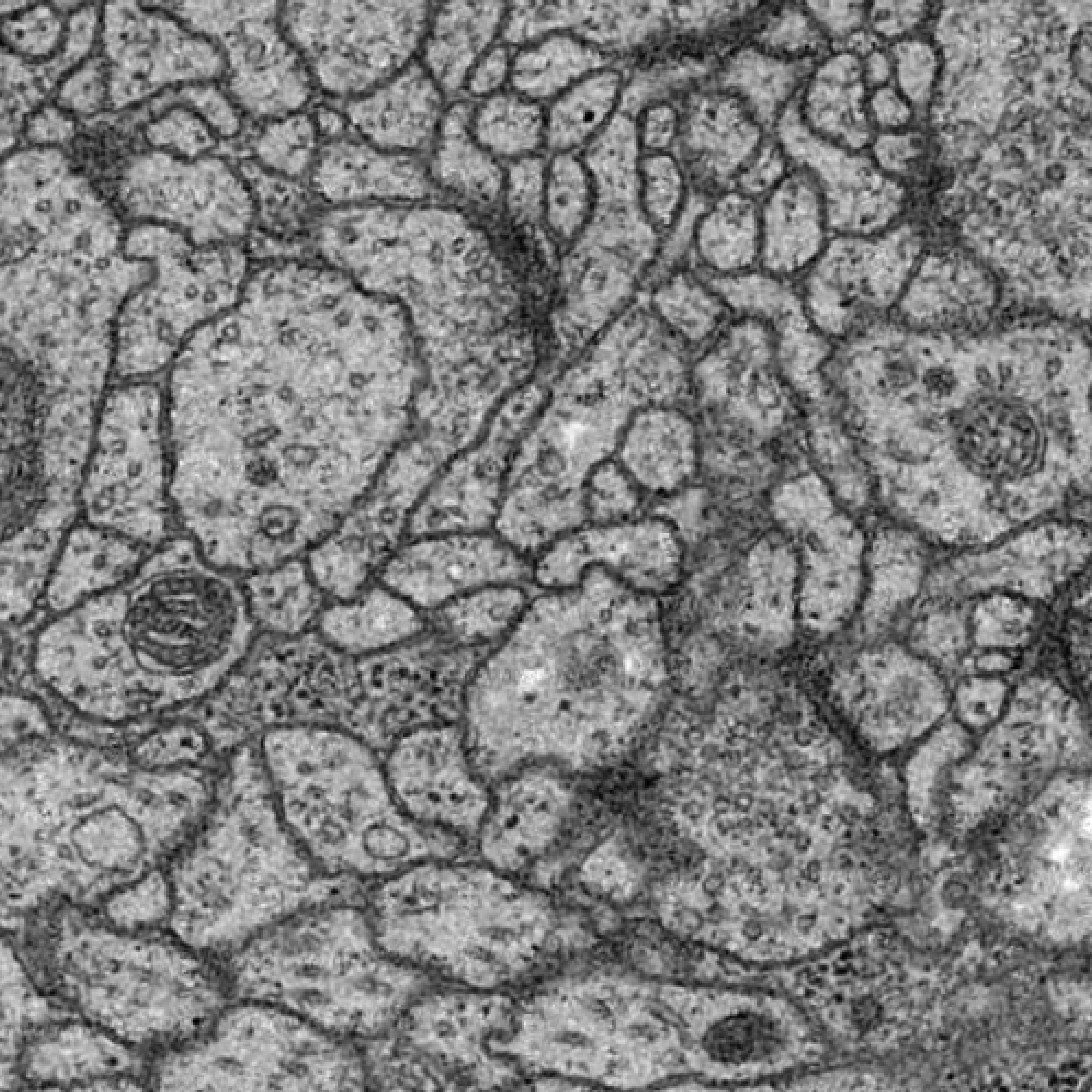}
  \end{subfigure}
  \begin{subfigure}[b]{0.24\textwidth}
    \includegraphics[width=\linewidth]{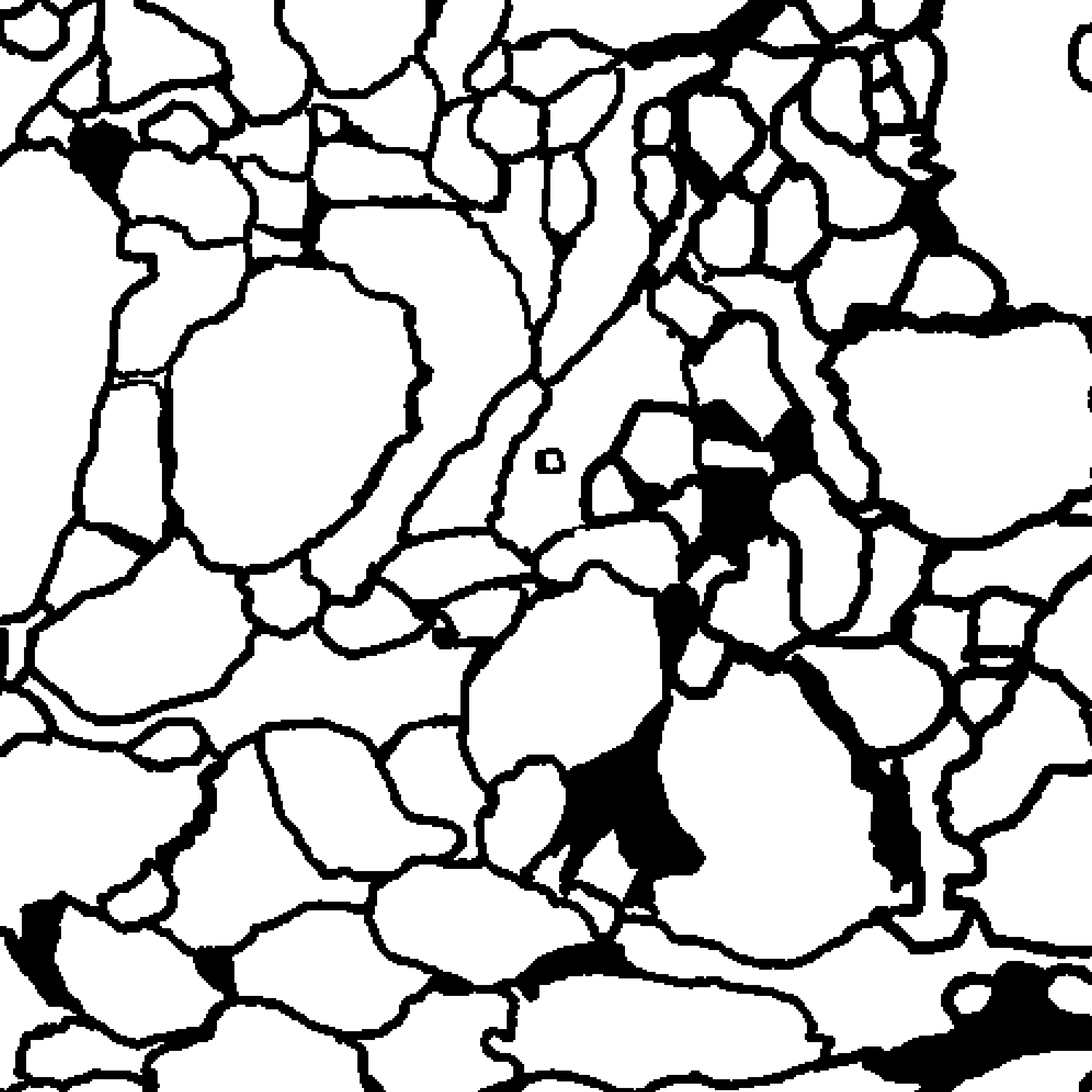}
  \end{subfigure}
  \begin{subfigure}[b]{0.24\textwidth}
    \includegraphics[width=\linewidth]{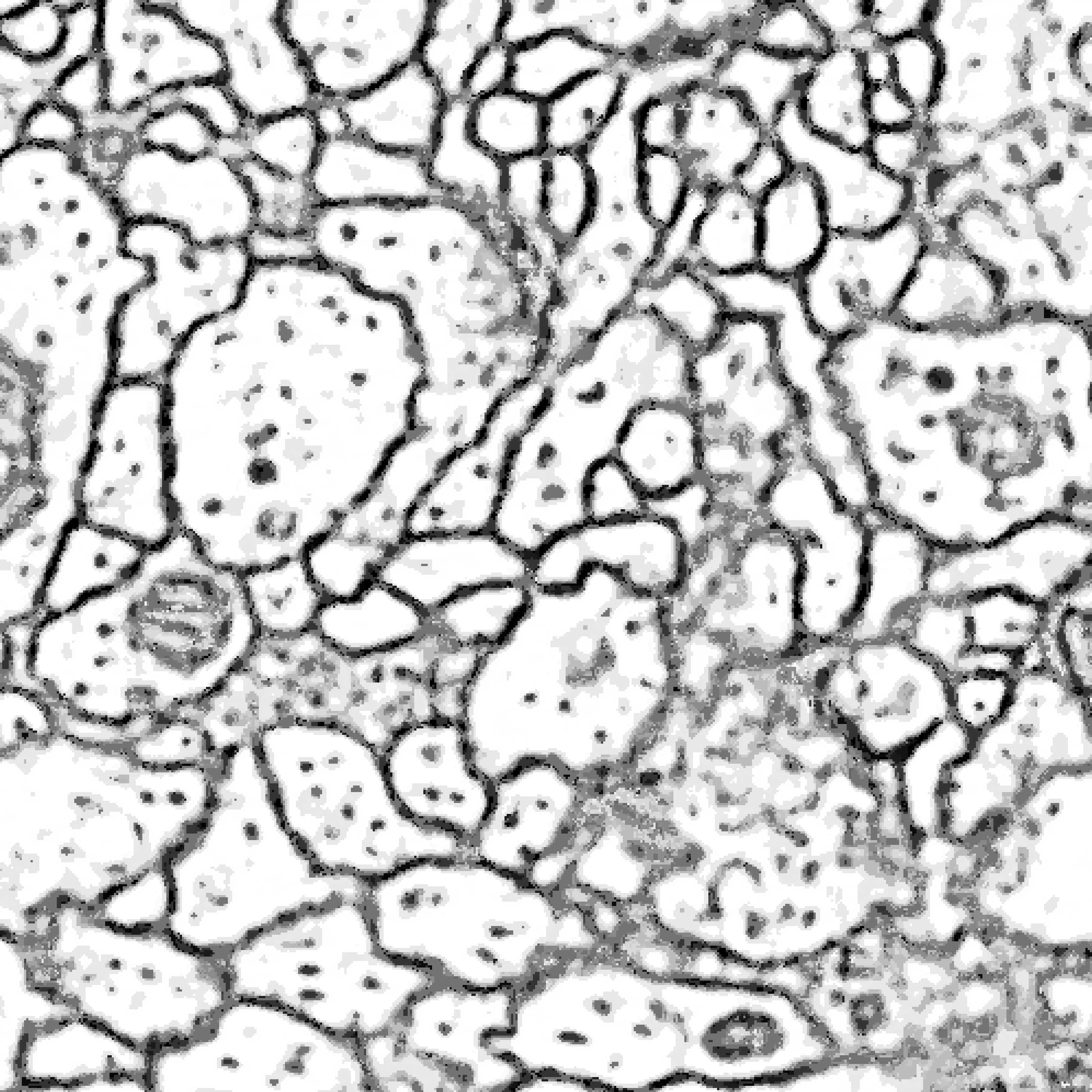}
  \end{subfigure}
  \begin{subfigure}[b]{0.24\textwidth}
    \includegraphics[width=\linewidth]{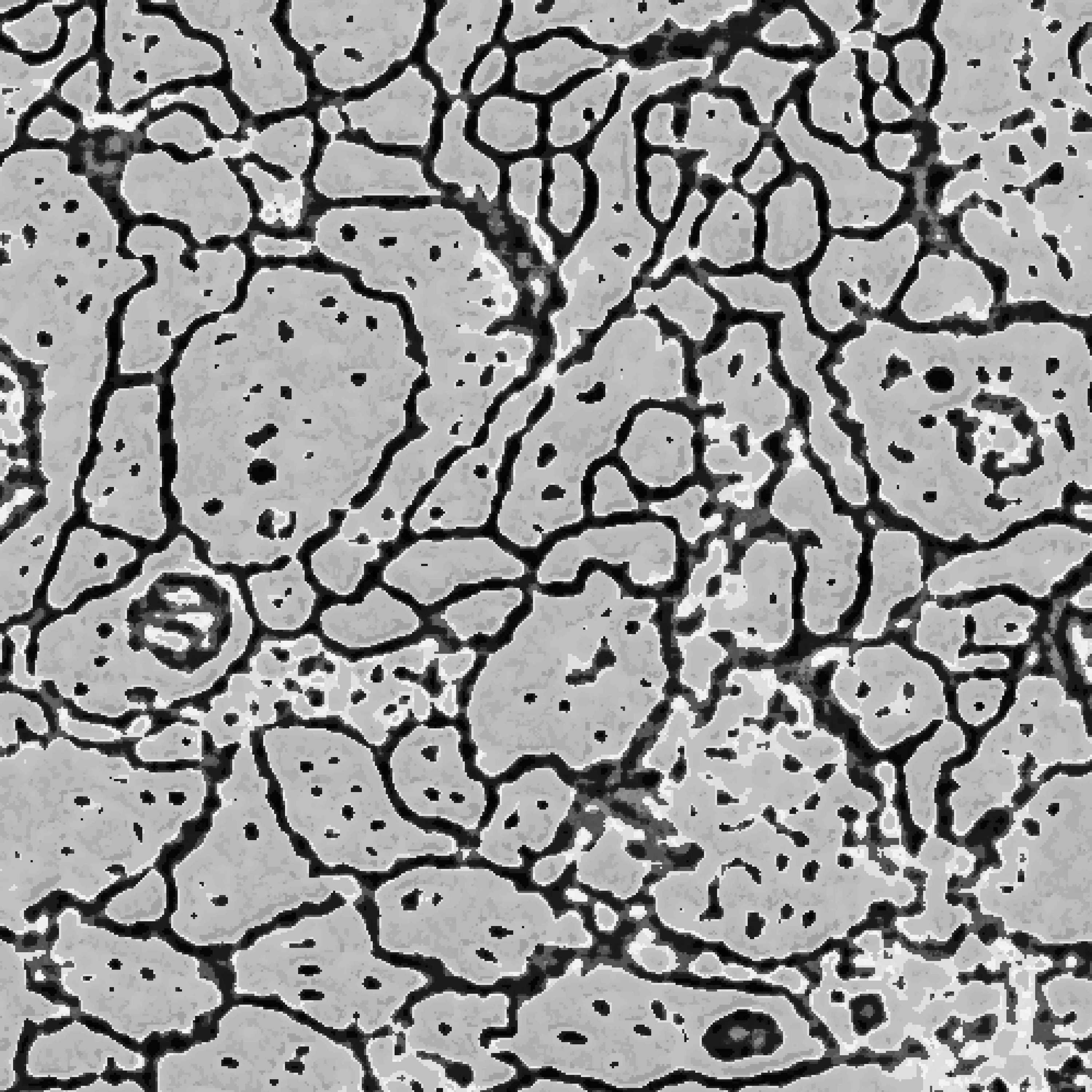}
  \end{subfigure}
  \begin{subfigure}{0.24\textwidth}
    \caption{input}
    \label{fig:ISBIinput}
  \end{subfigure}
  \begin{subfigure}{0.24\textwidth}
    \caption{ground truth}
    \label{fig:ISBIgt}
  \end{subfigure}
  \begin{subfigure}{0.24\textwidth}
    \caption{prediction}
    \label{fig:ISBIpred}
  \end{subfigure}
  \begin{subfigure}{0.24\textwidth}
    \caption{leaves}
    \label{fig:ISBIleaves}
  \end{subfigure}
  \caption{Visualization of results of oblique decision trees on the ISBI binary segmentation 
           dataset.
           Column \subref{fig:ISBIinput} shows the input image and column \subref{fig:ISBIgt}
           the corresponding groundtruth labels.
           Column \subref{fig:ISBIpred} illustrates the probability estimate predicted by the
           oblique decision tree. Darker means higher probability for class ``membrane''.
           Column \subref{fig:ISBIleaves} show the leaf affiliation in the oblique decision tree.
           The leaves from left to right are equidistantly assigned to 
           grayscale values from black to white.}
  \label{fig:ISBI}
\end{figure*}
  We test the applicability of the proposed decision tree algorithm for image 
  segmentation on the ISBI challenge dataset \cite{ISBI}.
  This image partitioning benchmark comprises serial section Transmission
  Electron Microscopy images and binary annotations of neurons and membranes
  (figure~\ref{fig:ISBIinput}).
  \\
  For every pixel, we provide a sliding window around the current pixel as input
  features to the oblique decision tree.
  Consequently, the learned parameters at each split node can be regarded as a
  spatial kernel.
  We learn an oblique decision tree of depth 8 with a maximum of 256 leaves
  with our greedy EM algorithm.
  We use the default parameters as described in section~\ref{sec:EXPaccuracy}
  and train each split for 40 epochs.
  
  \textbf{Results.}
  Figure~\ref{fig:ISBI} shows a sample image of the input, the groundtruth
  labels, the predicted probability of our oblique decision tree and the 
  color-coded leaf affiliation.
  The visualization of the prediction shows pixels more likely to be of
  class ``membrane'' in darker color.
  In the color-coded leaf affiliation, each grayscale value represents a leaf
  in the oblique decision tree.
  Darker pixels have reached a leaf further on the left side of the 
  decision tree.
  \\
  In the prediction most of the membranes are correctly identified.
  However, many mitochondria are falsely classified as membrane.
  Interestingly, the leaf affiliation is fairly regular, implying that
  most adjacent pixels are routed to the same leaf, in spite of some variation
  in appearance.
  The leaf affiliation could also be provided as additional feature
  in order to stack classifiers.

\subsection{CNN split features} \label{sec:EXPCNN}
  In a preliminary experiment, we test the effectiveness of Convolutional
  Neural Networks as split features on \textit{MNIST}.
  At each split we trained a very simple CNN of the following architecture:
  Convolution \(5\times5\) kernel @ 3 output channels
  \(\rightarrow\) Max Pool \(2\times2\)
  \(\rightarrow\) ReLU
  \(\rightarrow\) Convolution \(5\times5\) @ 6 
  \(\rightarrow\) Max Pool \(2\times2\)
  \(\rightarrow\) ReLU
  \(\rightarrow\) Fully connected layer \(96\times50\)
  \(\rightarrow\) ReLU
  \(\rightarrow\) Fully connected layer \(50\times1\).
  The final scalar output is the split feature, which is the input to the
  split function.
  \\
  Again, we train greedily to initialize the tree, however we split nodes in
  a best-first manner, based on highest information gain.
  As a result, the trees can be fairly unbalanced despite impure leaves.
  We now choose to stop at a maximum of 10 leaves, 
  as we aim to increase interpretability and efficiency by having one expert
  leaf per class.
  
  \textbf{Results.}
  In this setting, we achieve a test accuracy of \(0.982 \pm 0.003\)
  deterministic evaluation of nodes.
  This model provides interesting benefits in interpretability and efficiency,
  which are the main advantages of decision trees.
  When a sample was misclassified it is straightforward to find the split node
  that is responsible for the error.
  This offers interpretability as well as the possibility to improve the
  overall model.
  Other methods, such as \mbox{\emph{OneVsOne}} or \mbox{\emph{OneVsRest}} multi-class 
  approaches, provide similar interpretability, however at a much higher
  cost at test time.
  This is due to the fact that in a binary decision tree with \(K\) leaves,\ie
  a leaf for each class, it is sufficient to evaluate \(\mathcal{O}(\log K)\)
  split nodes.
  In \mbox{\emph{OneVsOne}} and \mbox{\emph{OneVsAll}} it is necessary to 
  evaluate \(K (K-1) / 2\) and respectively \(K\) different classifiers at
  test time.

\section{Conclusion}
  We have presented a new approach to train deterministic decision trees with
gradient-based optimization in an end-to-end manner.
We show that this approach outperforms previous algorithms for oblique 
decision trees.
The approach is not restricted in the complexity of the split features and
we have provided preliminary evidence of the effectiveness of more complex
split features, such as convolutional neural networks.
Moreover, our approach allows imposing additional regularization constraints
on the learned split features.
We have demonstrated these capabilities by visualizing spatially 
regularized parameters on image processing datasets.
The overall approach provides high flexibility and the potential for 
accurate models that maintain interpretability and efficiency due to the
conditional data flow.

{\small
\bibliographystyle{ieee}
\bibliography{literature}
}

\end{document}


\title{Supplementary Material to\\End-to-end Learning of Deterministic Decision Trees}

\author{
  Thomas Hehn
  \quad\quad\quad\quad\quad
  Fred A. Hamprecht\\
  HCI/IWR, Heidelberg University\\
  {\tt\small \{thomas.hehn,fred.hamprecht\}@iwr.uni-heidelberg.de}
}

\maketitle

\begin{figure*}[!ht]
  \centering
  \begin{tikzpicture}
    \node (MNISTtest)  {\includegraphics[width=0.24\linewidth]{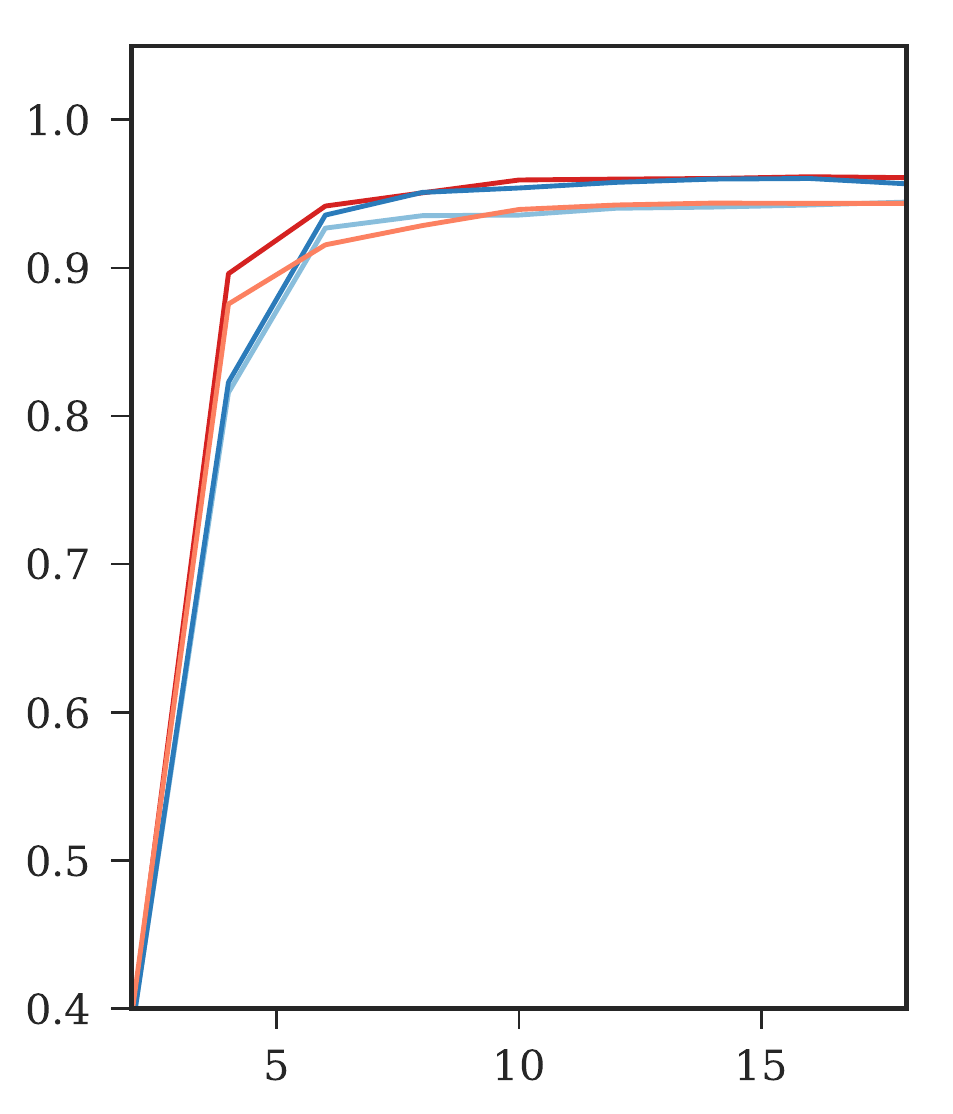}};
    \node[right=of MNISTtest, node distance=0cm, xshift=-1.2cm]
          (SensITtest)  {\includegraphics[width=0.24\linewidth]{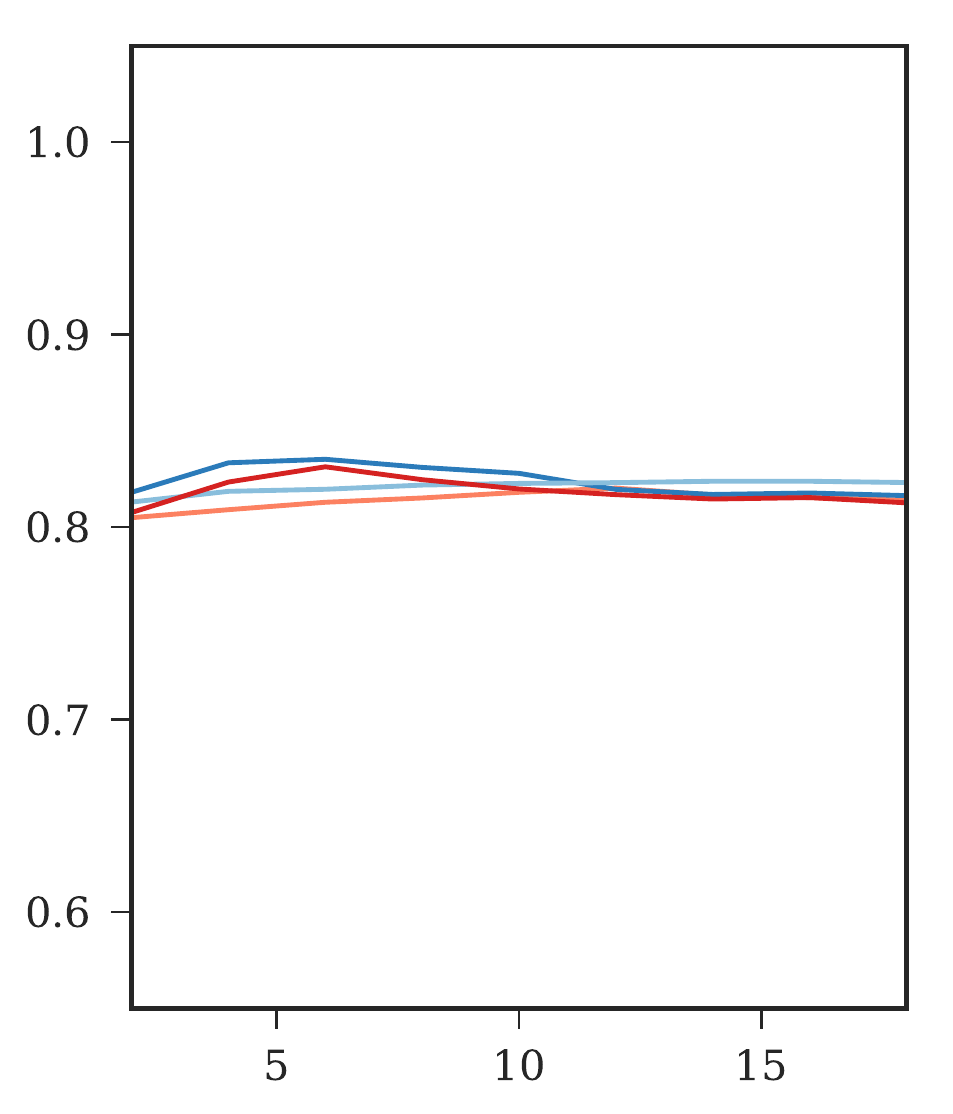}};
    \node[right=of SensITtest, node distance=0cm, xshift=-1.2cm]
          (Connect4test)  {\includegraphics[width=0.24\linewidth]{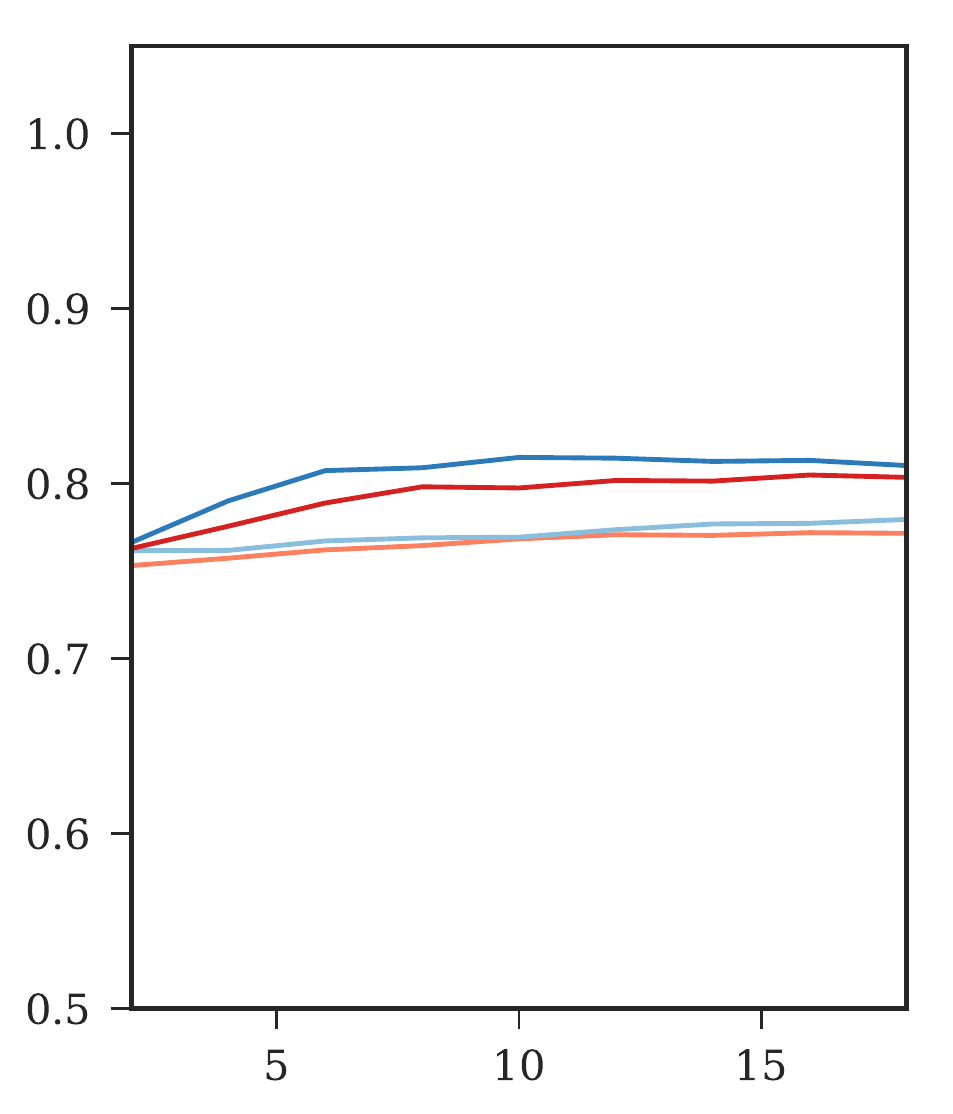}};
    \node[right=of Connect4test, node distance=0cm, xshift=-1.2cm]
          (Proteintest)  {\includegraphics[width=0.24\linewidth]{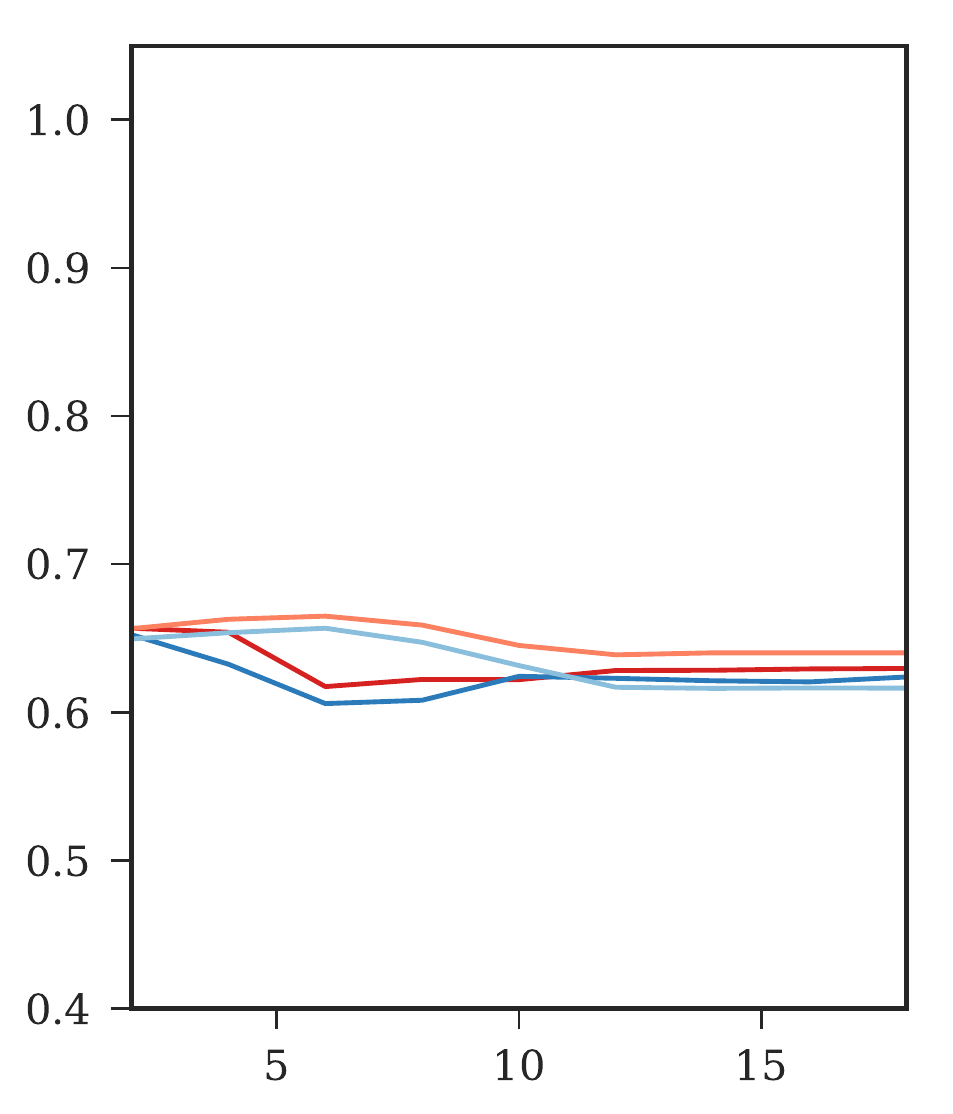}};
    \node[below=of MNISTtest, node distance=0cm, yshift=1.2cm] (MNISTtrain)  {\includegraphics[width=0.24\linewidth]{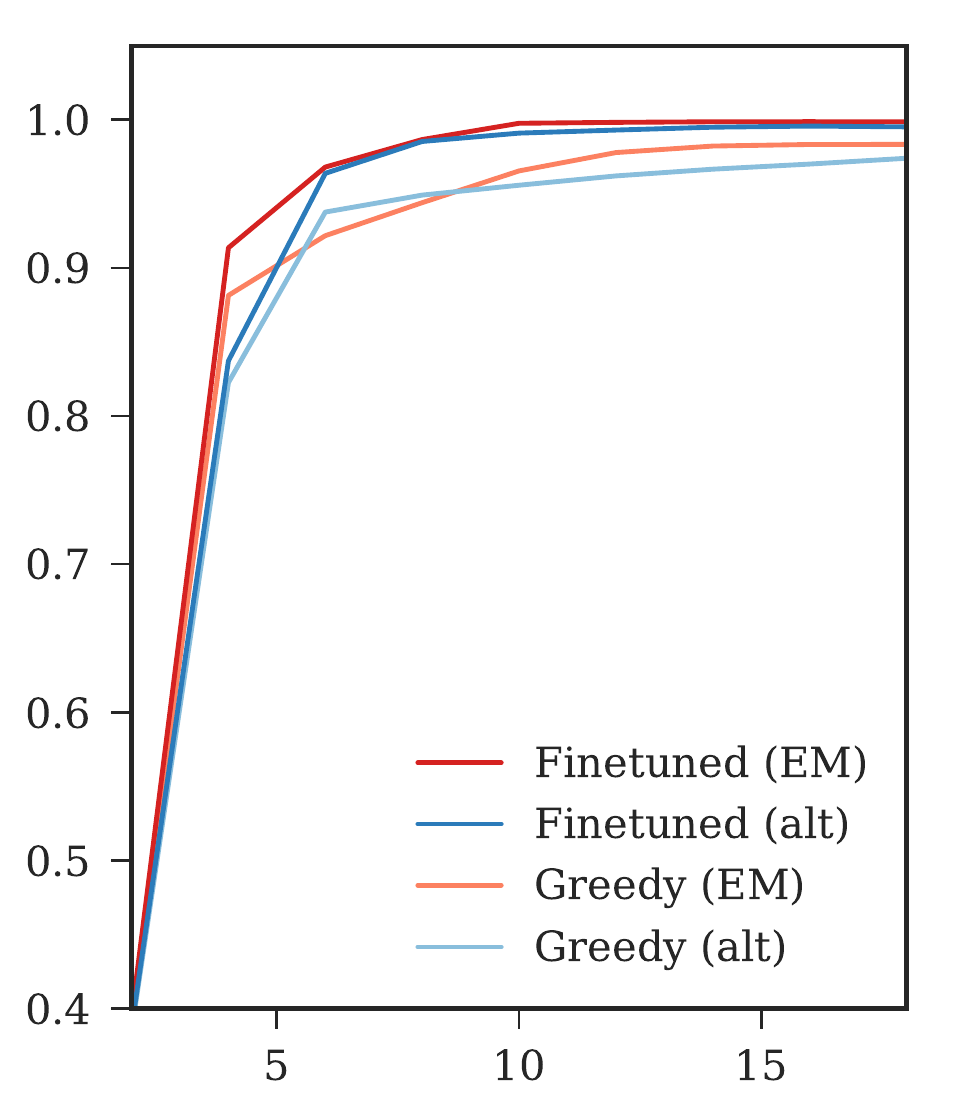}};
    \node[right=of MNISTtrain, node distance=0cm, xshift=-1.2cm]
          (SensITtrain)  {\includegraphics[width=0.24\linewidth]{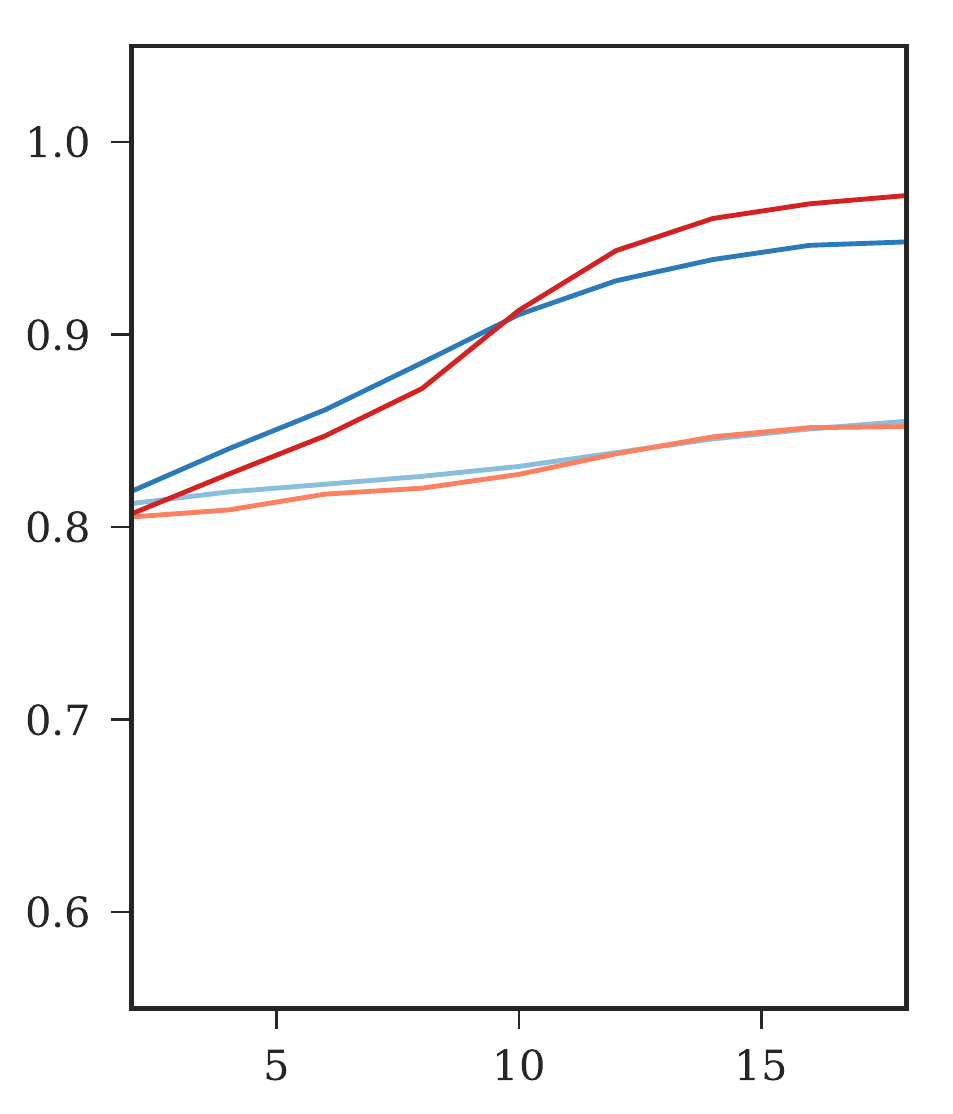}};
    \node[right=of SensITtrain, node distance=0cm, xshift=-1.2cm]
          (Connect4train)  {\includegraphics[width=0.24\linewidth]{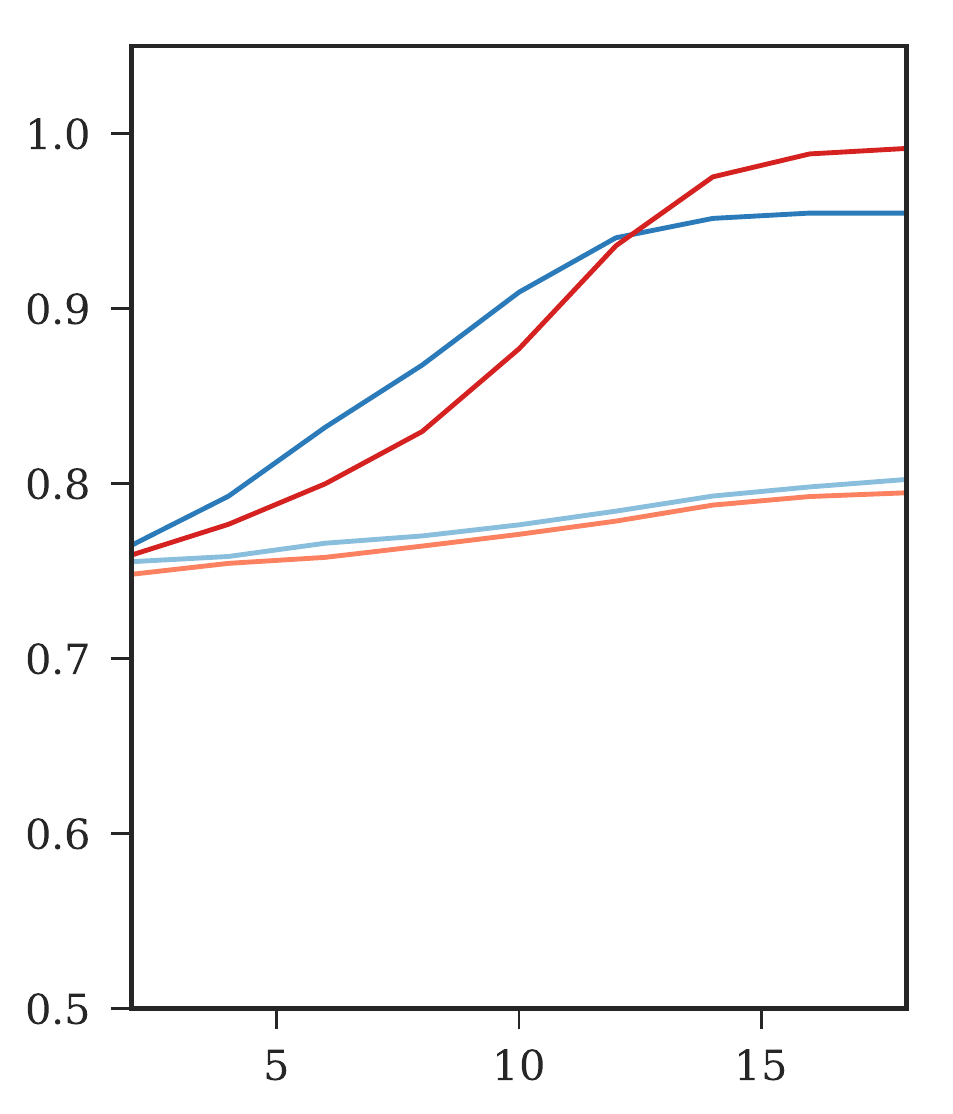}};
    \node[right=of Connect4train, node distance=0cm, xshift=-1.2cm]
          (Proteintrain)  {\includegraphics[width=0.24\linewidth]{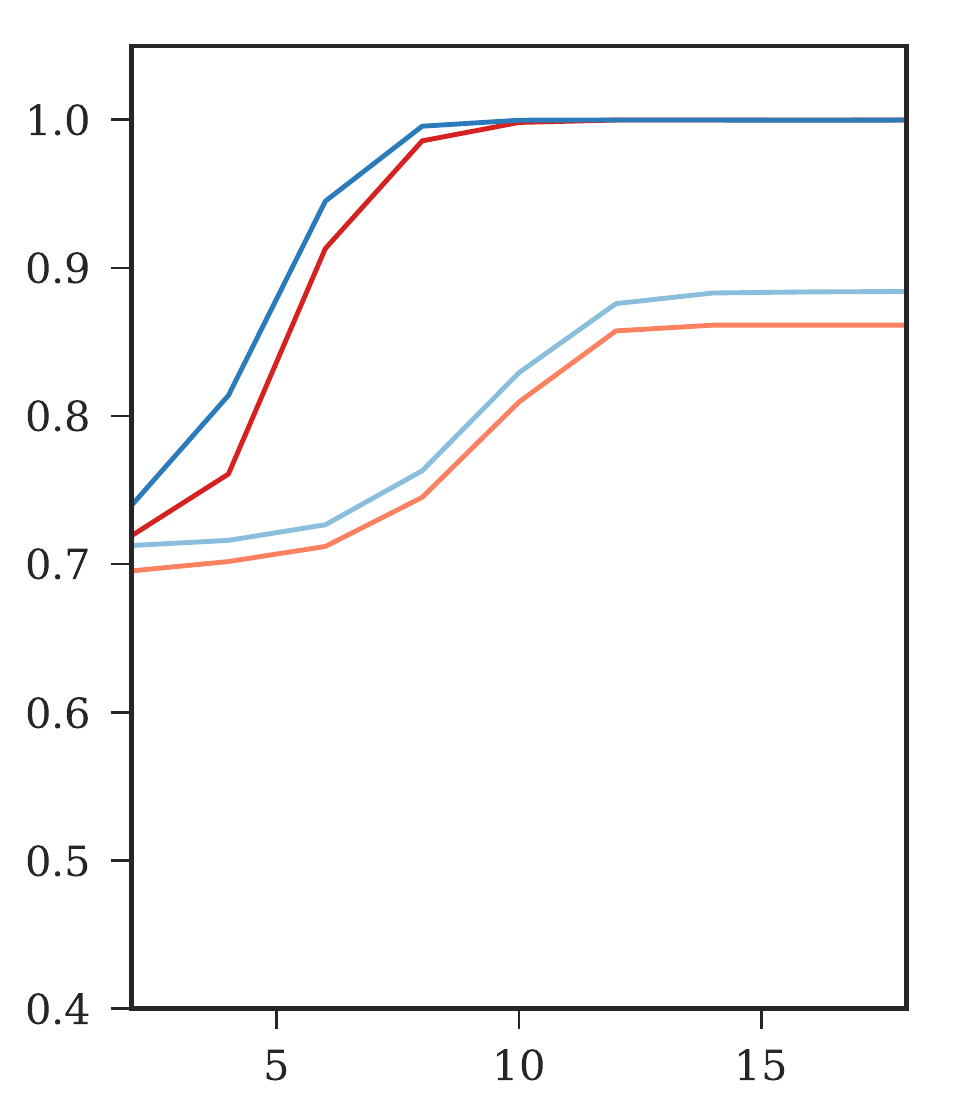}};

    \node[below=of MNISTtrain, node distance=0cm, font=\small,
          xshift=0.3cm, yshift=1.3cm] {Tree depth};
    \node[below=of SensITtrain, node distance=0cm, font=\small,
          xshift=0.3cm, yshift=1.3cm] {Tree depth};
    \node[below=of Connect4train, node distance=0cm, font=\small,
          xshift=0.3cm, yshift=1.3cm] {Tree depth};
    \node[below=of Proteintrain, node distance=0cm, font=\small,
          xshift=0.3cm, yshift=1.3cm] {Tree depth};

    \node[left=of MNISTtest, node distance=0cm, rotate=90, font=\small,
          anchor=center,yshift=-1.0cm] {Test accuracy};
    \node[left=of MNISTtrain, node distance=0cm, rotate=90, font=\small,
          anchor=center,yshift=-1.0cm] {Training accuracy};
  \end{tikzpicture}
  \begin{subfigure}[b]{0.24\textwidth}
    \caption{MNIST}
  \end{subfigure}
  \begin{subfigure}[b]{0.24\textwidth}
    \caption{SensIT}
  \end{subfigure}
  \begin{subfigure}[b]{0.24\textwidth}
    \caption{Connect4}
  \end{subfigure}
  \begin{subfigure}[b]{0.24\textwidth}
    \caption{Protein}
  \end{subfigure}
  \caption{Performance of alternating optimization (\textit{alt}) compared to 
           the EM optimization (\textit{EM}) presented in the main paper.
           Accuracy of oblique decision trees at different maximum depth
           on test and training sets is compared.
           Maximum tree depth is varied from 2 to 18 with step size 2.
           At each depth, both algorithms (\textit{alt} and \textit{EM}) are 
           used to learn respectively an initial \textit{greedy} tree and 
           \textit{finetune} the initial tree.
           }
  \label{fig:SUPresults}
\end{figure*}
\section{Alternating optimization} \label{sec:altopt}
  Similarly to the EM-algorithm presented in section~3.3 of the main paper, the
  alternating strategy aims to
  optimize the split parameters and the leaf prediction parameters separately.
  However, instead of computing the exact leaf predictions, we approximate
  them using a deterministic decision tree.
  \\
  Let \(N_{\ell,k}\) be the number of training samples in leaf \(\ell\) with
  class \(k\) and \(N_{\ell} = \sum_{k=1}^K N_{\ell,k}\) denotes the sum of 
  all training samples in leaf \(\ell\).
  The optimal leaf predictions are
  \begin{equation}
    (\bm{\pi}_\ell)_k = \frac{N_{\ell,k}}{N_{\ell}}.
    \label{eq:discreteleaves}
  \end{equation}
  Analogously to the EM-algorithm in the main paper, we use Adam~\cite{Kingma2015}
  to optimize the split parameters.
  The log-likelihood objective given the current estimate of the
  leaf predictions is:
  \begin{equation}
    \max_{\bm{\theta}_{\bm{\beta}}}
    \sum_{n = 1}^N \log \left(
    \sum_{\ell=1}^L 
    (\bm{\pi}_\ell)_y
    \mu_\ell(\bm{x};\sigma_\gamma,\bm{\theta}_{\bm{\beta}})
    \right).
  \end{equation}
  This algorithm is applicable to optimize an entire tree, as well as to 
  optimize tree stumps as required for the greedy structure learning.
  \subsection{Experiments}
  We compare this algorithm to the results of our EM algorithm for oblique
  decision trees reported in section~4.1 of the main paper.
  The hyperparameter tuning and the experimental setup are done in the same
  way as for the EM algorithm.
  Figure~\ref{fig:SUPresults} illustrates the results of the experiments.
  The results show that both algorithms perform equally well in terms of accuracy.

\section{Visualization of oblique decision trees}
\begin{sidewaysfigure*} 
  \centering
  \includegraphics[width=\linewidth]{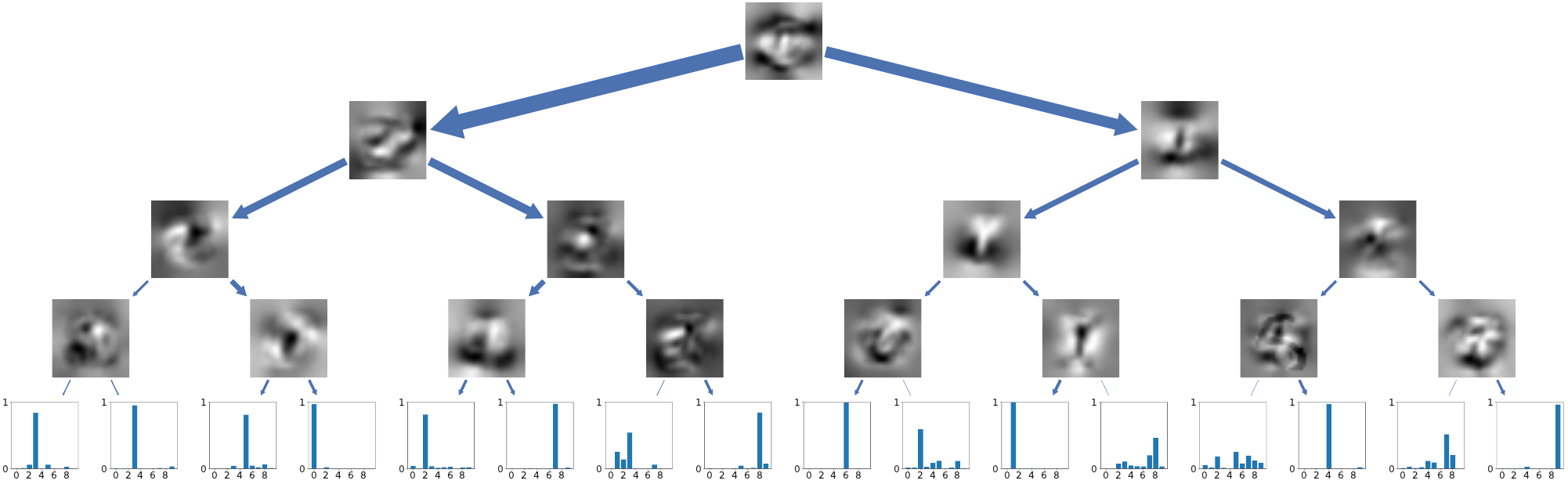}
  \caption{Visualization of an oblique decision tree learned on 
           \textit{MNIST}~\cite{LeCun1998} with spatial regularization.
           The split parameters are visualized as in section~4.2 of the
           main paper.
           As a rule of thumb: The better the input image matches the
           parameter image, the more likely the sample will go to the
           right child.
           If the input image better resembles the negative parameter
           image, the sample will go to the left.
           The thickness of an arrow indicates the number of training samples 
           following the path, when the decision tree is evaluated
           deterministically.
           Leaf predictions are visualized as bar plots, where the 
           \textit{x-axis} denotes classes 0 to 9 and the \textit{y-axis}
           corresponds to the probability of each class.
           Axis labels were omitted due to space restrictions.
           }
  \label{fig:SUPtreeMNIST}
\end{sidewaysfigure*}

\begin{sidewaysfigure*} 
  \centering
  \includegraphics[width=\linewidth]{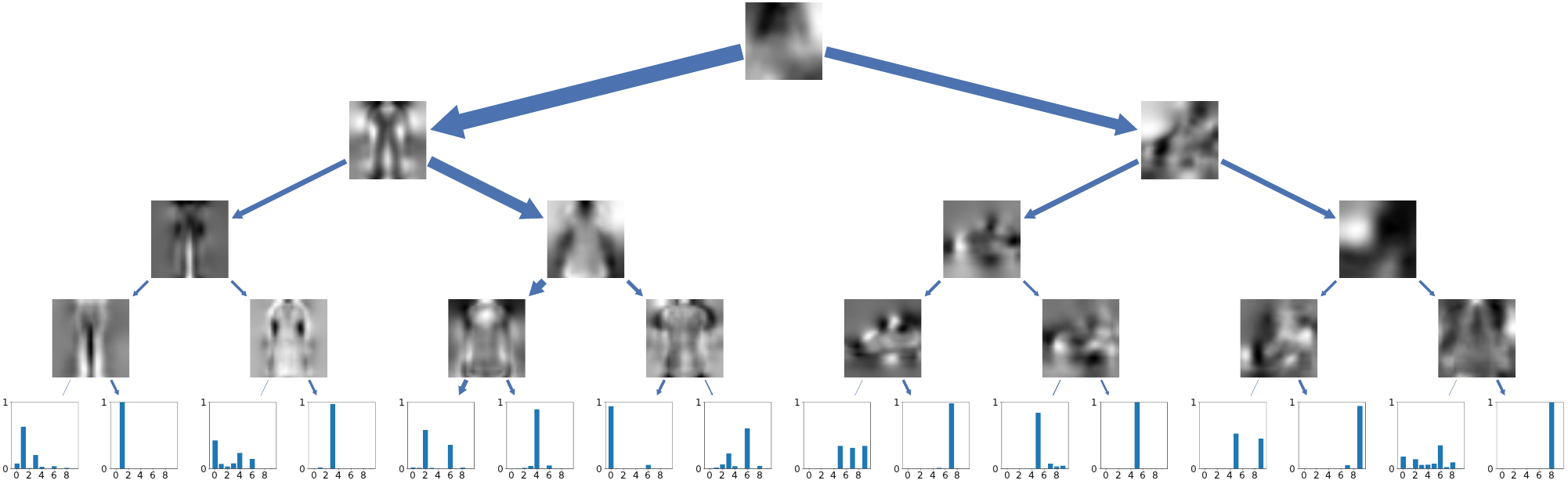}
  \caption{Visualization of an oblique decision tree learned on 
           \textit{FashionMNIST}~\cite{FashionMNIST} with spatial regularization.
           The split parameters are visualized as in section~4.2 of the
           main paper.
           As a rule of thumb: The better the input image matches the
           parameter image, the more likely the sample will go to the
           right child.
           If the input image better resembles the negative parameter
           image, the sample will go to the left.
           The thickness of an arrow indicates the number of training samples 
           following the path when the decision tree is evaluated
           deterministically.
           Leaf predictions are visualized as bar plots. 
           The \textit{x-axis} denotes
           classes (
           0: T-shirt/top, 
           1: Trouser, 
           2: Pullover, 
           3: Dress, 
           4: Coat, 
           5: Sandal, 
           6: Shirt, 
           7: Sneaker, 
           8: Bag, 
           9: Ankle boot)
           and the \textit{y-axis} corresponds to the probability of
           each class.
           Axis labels were omitted due to space restrictions.
           }
  \label{fig:SUPtreeFashionMNIST}
\end{sidewaysfigure*}

  Figures~\ref{fig:SUPtreeMNIST} and \ref{fig:SUPtreeFashionMNIST} 
  show visualizations of entire oblique decision trees of depth 4 trained
  on \textit{MNIST}~\cite{LeCun1998} and \mbox{\textit{FashionMNIST}}~\cite{FashionMNIST}.
  Both trees were trained with the EM algorithm and spatial regularization to
  obtain smooth visualizations.
  Leaf predictions are visualized as bar plots.
  The split parameters are visualized as in section~4.2 of the main paper.
  Intuitively, the better the input image matches the parameter image,
  the more likely it is to go to the right.
  Likewise, the better the input images resembles the negative parameter
  image, the more likely it is to go the left child.
  How many training samples follow a certain path is indicated by the
  thickness of the arrows.
  Thicker arrows mean more training samples follow the path when the
  decision tree is deterministic.

  Some split parameters of the decision tree trained on \textit{MNIST}
  (figure~\ref{fig:SUPtreeMNIST}) reveal interesting structures.
  At depth 3 (row 4), the fifth split node from the left shows a dark silhouette
  of the number ``6''.
  Accordingly, the left child predicts class ``6''.
  In the same manner, the parameters of the sixth split node at depth 3 expose a dark 
  stroke in the center surrounded by brighter pixels.
  This is used to distinguish class ``1'', which is predicted by the left child.
  \\
  The decision tree in figure~\ref{fig:SUPtreeFashionMNIST} was trained on
  the \mbox{\textit{FashionMNIST}} dataset~\cite{FashionMNIST},
  a dataset comprising \(28\times28\) grayscale images of fashion products.
  These fashion products can be discovered in the learned split parameters.
  The split parameters of the first split at depth 3 (row 4) show bright
  trousers and its right child predicts the class ``trousers''.
  The same holds for the second split at depth 3 showing a bright silhouette
  of a dress and its right child which predicts ``dress''.
  The parameters of the third split at depth 3 reveal some kind of upper body
  clothes, but it is difficult to determine the kind.
  Yet, these parameters separate samples of class ``pullover'' and
  ``shirt'' (left child) from class ``coat'' (right child).
  Similarly, the fifth split node also reveals only a slight silhouette of
  a shoe.
  Still, its right child is certain predicting ``sneaker'',
  whereas the left child is rather indecisively predicting ``sandals'',
  ``sneaker'' and ``ankle boots''.

  The decision tree illustrations reveal the internal decisions being made
  to reach a final prediction and provide a useful tool to interpret
  our model.
{\small
\bibliographystyle{ieee}
\bibliography{literature}
}